\documentclass[conference]{IEEEtran}
\IEEEoverridecommandlockouts
\usepackage{cite}
\usepackage{hyperref}
 \hypersetup{
	colorlinks=true,
	linkcolor=black,
	filecolor=black,      
	urlcolor=black,
	citecolor=black,
}
\usepackage{amsmath,amssymb,amsfonts}
\usepackage{graphicx}
\usepackage{textcomp}
\usepackage{algorithm}
\usepackage{algorithmic}
\usepackage{balance}

\floatname{algorithm}{Algorithm}
\def\BibTeX{{\rm B\kern-.05em{\sc i\kern-.025em b}\kern-.08em
		T\kern-.1667em\lower.7ex\hbox{E}\kern-.125emX}}

\usepackage{caption} %many figures in one figure (note subfigure and subfig are deprecated)
\usepackage{subcaption} %many figures in one figure (note subfigure and subfig are deprecated)
\usepackage{multirow}
\usepackage{enumerate}

\usepackage{amsmath}
\usepackage{amsfonts}
\usepackage{mathtools}

\usepackage{amssymb}%............................ AMS Symbol fonts
\usepackage{amsthm}%............................. Theorems
\usepackage[T1]{fontenc}%........................ Type 1 fonts
\usepackage{bm}%................................. Bold math symbols (after fonts)
\usepackage{tikz}

\newcommand\copyrighttext{% 
	\footnotesize \textcopyright 2020 IEEE. Personal use of this material is permitted. Permission from IEEE must be obtained for all other uses, in any current or future media, including reprinting/republishing this material for advertising or promotional purposes, creating new collective works, for resale or redistribution to servers or lists, or reuse of any copyrighted component of this work in other works. }

\newcommand\copyrightnotice{%
	\begin{tikzpicture}[remember picture,overlay]\node[anchor=south,yshift=10pt] at (current page.south) {\fbox{\parbox{\dimexpr\textwidth-\fboxsep-\fboxrule\relax}{\copyrighttext}}};
	\end{tikzpicture}%
}

\begin{document}

\title{Performance-Agnostic Fusion of Probabilistic Classifier Outputs\\
{\footnotesize \textsuperscript{*}}
\thanks{JFM and SWU are members of the AIMS Research Centre and thank the Centre for its support.
	The AIMS Research Centre receives funding support from the National Research Foundation, South Africa, which is also acknowledged with thanks.
	JFM thanks the DST-CSIR HCD Inter-Bursary scheme for PhD funding.}
}

\author{\IEEEauthorblockN{Jordan F. Masakuna}
	\IEEEauthorblockA{
		\textit{Computer Science Division} \\
		\textit{Stellenbosch University}\\
		Stellenbosch, South Africa}
	\IEEEauthorblockA{\textit{and African Institute for Mathematical Sciences} \\
		%\textit{name of organization (of Aff.)}\\
		Cape Town, South Africa \\
		jordan@aims.ac.za}
	\and
	\IEEEauthorblockN{Simukai W. Utete}
	\IEEEauthorblockA{\textit{African Institute for Mathematical Sciences} \\
		%\textit{name of organization (of Aff.)}\\
		Cape Town, South Africa \\
		simukai@aims.ac.za}
	\and
	\IEEEauthorblockN{Steve Kroon}
	\IEEEauthorblockA{\textit{Computer Science Division} \\
		\textit{Stellenbosch University}\\
		Stellenbosch, South Africa \\
		kroon@sun.ac.za}
}

\maketitle
\copyrightnotice
\begin{abstract}
We propose a method for combining probabilistic outputs of classifiers to make a single consensus class prediction when no further information about the individual classifiers is available, beyond that they have been trained for the same task.
%This paper proposes a message-passing algorithm to combine black-box classifiers. Here agents are equipped with classifiers to perform a classification task. 
%There is a single class of interest which might be inconsistently described by the group of classifiers.
The lack of relevant prior information rules out typical applications of Bayesian or Dempster-Shafer methods, and the default approach here would be methods based on the principle of indifference, such as the sum or product rule, which essentially weight all classifiers equally.
In contrast, our approach considers the diversity between the outputs of the various classifiers, iteratively updating predictions based on their correspondence with other predictions until the predictions converge to a consensus decision.
%To tackle inconsistency in fusion of classifiers, diversity of classifiers needs to be taken into account \cite{finkelstein1994inconsistency}. 
The intuition behind this approach is that classifiers trained for the same task should typically exhibit regularities in their outputs on a new task; the predictions of classifiers which differ significantly from those of others are thus given less credence using our approach.
The approach implicitly assumes a symmetric loss function, in that the relative cost of various prediction errors are not taken into account.
%The proposed model extracts information from outputs of the classifiers by looking at them in relation to each other. 
%Bayesian and Dempster-Shafer methods, amongst others, can handle such situations.
%We consider situations where these classifiers are provided only as black boxes: separately pre-trained classifiers without quantitative information about their performance.
% 
%However, the methods mentioned above would require evidence to distinguish hypotheses.
%
%However, methods based upon the principle of indifference weight classifiers' outputs equally.
%Our objective is to combine black-box classifiers using a message-passing algorithm to achieve consensus, with agents updating their beliefs based on messages received from other agents until consensus is achieved. 
Performance of the model is demonstrated on different benchmark datasets.
Our proposed method works well in situations where accuracy is the performance metric; however, it does not output calibrated probabilities, so it is not suitable in situations where such probabilities are required for further processing.
%When distributions converge, our proposed method assigns high probability (of almost one) to the consensus class, and low probability (of almost zero) to other classes.
%So when there is misclassification, the likelihood-like loss gets high.
\end{abstract}

\begin{IEEEkeywords}
Classifier fusion, ensemble methods, consensus theory
\end{IEEEkeywords}

\section{Introduction}
%Applications of the problem of decision fusion~\cite{kittler1998combining, castanedo2013review} include rating and ranking \cite{langville2012s}, classification \cite{kuncheva2002switching} and remote sensing \cite{utete1999voting}.
%
%The problem of decision fusion arises when outputs from various agents must be combined. 

In classifying an observation, the members of an ensemble of (probabilistic) classifiers will typically disagree, with individual classifiers sometimes producing markedly different predictions for the same observation.
Leveraging such disagreement can be the very basis of improved overall performance, for example in boosting, where different subsets of classifiers might perform better in different settings~\cite{schapire2003boosting}.
Similar predictions are typically easy to handle when a single class prediction must be made; however, it is trickier to reach consensus decisions for markedly different predictions~\cite{finkelstein1994inconsistency}.

We consider the problem of fusing probabilistic classifier predictions to reach a consensus decision, i.e. a decision all members agree to support in the best interest of the whole group~\cite{lynch2008decentralized}. 
%It is possible, especially with the application of transfer learning \cite{pan2009survey} where a classifier trained for a classification task (e.g. recognition of cars) can be used for another classification task (e.g. recognition of trucks), that one combines classifiers  trained  for different classification tasks where classifiers outputs might not be similar.
The key assumption we make is that the various classifiers have been trained to perform the same classification task.
As a result, we would expect the classifier outputs to be quite similar, particularly in regions where they have fairly high confidence.
In regions where a classifier is not confident about class probabilities, it is more likely that its prediction deviates more notably from those of other classifiers.
%This might be applicable in robotics when sensor nodes of a robot perform classification tasks. A robot performing an object recognition task may possess multiple sensor nodes. Each node separately makes a decision about the type of an object based on information from its associated sensors. %In the application of mining, a sensor node should be able to recognise types of minerals.

%We consider classifier fusion problems where the aim is to combine outputs from multiple classifiers pre-trained on the same task to achieve a consensus, which improves over the individual classifier outputs.

%A variety of methods have been developed for using classifiers in ensembles. 
%
%One of the key motivations for ensemble techniques in machine learning is the no free lunch theorem \cite{wolpert1997no}, which states the impossibility of realising a universal best classification methodology. 
%
Various approaches have been developed~\cite{mangai2010survey} to take advantage of diverse classifier performance in different settings, including Bayesian methods~\cite{kim2012bayesian} and Dempster-Shafer~(DS) methods~\cite{fei2019novel}.
However, these approaches invariably make use of information about prior performance of the classifiers, such as the training or test error.
Such information about classifier performance is invaluable for high-performance classifier fusion, since it allows one to account explicitly for the quality of each classifier.
For DS methods, for instance, such information is used to form basic probability assignments of all subsets of classes~\cite{fei2019novel}. %(Basic probability assignments are required in Dempster-Shafer methods).
This paper considers the challenging situation where the classifiers are particularly opaque, more so than in the traditional black-box setting: we do not know what model each classifier uses, how or on what data the classifier was trained, what features it was trained with, its performance on any past data, or any of its predictions on any previous observations. 
%To our knowledge
%As far as we are aware, very little work has been done on this problem in quite a long time.

In this setting, the most natural approach would be to use the principle of indifference~\cite{keynes1921chapter}, treating all of the classifiers equally.
Prominent approaches based on this principle are the sum rule, the product rule, the majority vote rule~\cite{kittler1998combining1} and the Borda count rule~\cite{bordacount}.\footnote{The product rule corresponds to Bayesian decision fusion with a uniform prior over classifiers, assuming one of the classifiers is correct.}
The approach we present here relies on the assumption that simply because various classifiers were trained for the same task, we expect there to be regularities in their outputs on future observations, which we should be able to leverage to obtain improved fused decisions relative to approaches using the principle of indifference.

This work is inspired by the negotiation process, where stakeholders (classifiers) attempt to reach consensus decisions: stakeholders gradually update their position during the negotiations until consensus is reached.
Our approach is similarly an iterative process, in which initial classifier outputs are transformed to become more similar with the aim of reaching consensus.

\subsection*{\bf Contributions}
\label{sec:fusion_contribution}
%Information about the performance of classifiers in classifier fusion is very valuable. Such information permits a classifier fusion method to take into account the quality of the classifiers when combining classifiers' outputs. 
%In the absence of such information, it may seem that classifiers should all be treated equally.
This paper introduces a classifier fusion method which takes advantage of prediction similarities to enhance consensus decision-making in once-off classifier fusion.
While classifiers are not ultimately weighted equally, the approach is rooted in the principle of indifference, in that no external information is used to weight their relative importance.
% but which takes into account a sort of support of classifiers in the fusion process.
Rather, the fusion mechanism used leverages discrepancies between the predictions of the various classifiers on the observation of interest: fusion is essentially achieved by updating a probability vector for each classifier---initialized with the initial predictions---based on the discrepancies between them until they agree.

We empirically evaluate our proposed approach on a number of benchmark datasets from the UCI machine learning repository\cite{asuncion2007uci}, as well as one from the Columbia object image library\cite{nene1996columbia}.
We observe that the accuracy of our approach generally exceeds those of the sum, product, majority vote, and Borda count rules in a variety of settings.

\section{Background and Literature Review}
\label{chp:literature_classification}
While there is a vast literature on classifier fusion, the ``zero-knowledge'' situation we consider in this paper has received relatively little attention.
In general, the lack of knowledge about the various classifiers would lead one to consider approaches based on the principle of indifference.
Kittler et al. analyse and empirically compare various fusion approaches based on this principle, and note that the sum rule exhibited the best performance~\cite{kittler1998combining1}.
They attribute this to its lower sensitivity to estimation errors in individual classifier outputs.
%In \cite{kittler1998combining, kittler1998combining1},  Kittler et al. indicated that, in terms of performance, the sum rule is the best fusion rule based upon the principle of indifference: after conducting an analysis of error sensitivity (see Subsection \ref{sec:sensitivity_old}) and experimental investigations of the sum rule versus the product rule, then showed that the sum rule is much less affected by estimation errors. 

We next introduce the sum rule, product rule, majority vote rule and Borda count rule.
%after introducing some notation.
%We will compare our approach to the sum rule.
We wish to combine $m$ classifiers' predictions for an $l$-class classification task on a single input observation \textbf{$x_\text{in}$}.
%There are $m$ classifiers, and each one's output values are treated as probability vectors.
Specifically, the $j$-th classifier $f_j$ outputs a probability vector $d_j=f_j(x_\textbf{in})$ of length $l$. %For simplicity of notation, $f_j(x_\text{in})$ is denoted by $d_j$.
%
%, and let the classifiers' outputs for input $x_\textbf{in}$ be represented by the following output matrix,  \snote{Commenting out - seems to not be used, since it is not given a symbol?  Referenced in equations below, but matrix not really required.}
%\begin{equation}
%\label{eq:bmatrix}
%\begin{bmatrix}
%d_1\\
%\vdots\\
%d_m
%\end{bmatrix}\,,
%\end{equation}
%where $d_j=(d_{j1}, d_{j2}, \cdots, d_{jl})$.
% and $l$ denotes the number of class labels.

\paragraph{Sum rule}
\label{subsec:cam}
%Considering the classifiers' outputs in Equation \ref{eq:bmatrix},
The fused output using the sum rule is
\begin{equation}
\label{eq:argmax}
\frac{1}{m} \sum_{j=1}^m d_j=\frac{1}{m}\bigg( \sum_{j=1}^m d_{j1}, \sum_{j=1}^m d_{j2}, \cdots, \sum_{j=1}^m d_{jl}\bigg) \enspace .
\end{equation}

\paragraph{Product rule}
\label{subsec:prod}
%Considering the classifiers' outputs in Equation \ref{eq:bmatrix},
The fused output using the product rule is
\begin{equation}
\label{eq:argmax_prod}
\alpha \prod_{j=1}^m d_j \equiv \alpha\bigg( \prod_{j=1}^m d_{j1}, \prod_{j=1}^m d_{j2}, \cdots, \prod_{j=1}^m d_{jl}\bigg) \enspace ,
\end{equation} 
where $\alpha$ denotes a normalisation constant.
\paragraph{Majority vote rule}
\label{subsec:majority}
%Considering the classifiers' outputs in Equation \ref{eq:bmatrix},
Let $$
\Delta_{jk} = \begin{cases} 
1 & \text{if } d_{jk}=\max_{s=1}^l d_{js} \\
0      & \text{otherwise. }
\end{cases}
$$
The fused output using the majority vote rule is
\begin{equation}
\frac{1}{m}\bigg( \sum_{j=1}^m \Delta_{j1}, \sum_{j=1}^m \Delta_{j2}, \cdots, \sum_{j=1}^m \Delta_{jl}\bigg)\,.
\end{equation} 
\paragraph{Borda count rule} In the Borda count rule, classifiers rank classes in order of preference by giving each class a number of points corresponding to the number of classes ranked lower.
\label{subsec:majority2}
%Considering the classifiers' outputs in Equation \ref{eq:bmatrix},
Let 
$$
\beta_{jk} = l-|\{d_{ji}:d_{ji}>d_{jk}\lor(d_{ji}=d_{jk}\land i<k)\}|\,.
$$

The fused output using the Borda count rule is
\begin{equation}
\frac{2}{ml(l+1)}\bigg( \sum_{j=1}^m \beta_{j1}, \sum_{j=1}^m \beta_{j2}, \cdots, \sum_{j=1}^m \beta_{jl}\bigg)\,.
\end{equation} 
%
%Note  that, unlike the sum rule and product rule which outputs a calibrated probability distribution, the majority vote rule outputs a decision: the final class to assign to the object of interest. As described in Equation \ref{eq:argmax_majority}, the procedure for the majority vote rule simply  consists of counting  the  votes  received  for  each  class from  the  individual  classifiers.  The  class  which receives  the  largest  number  of  votes  is  then  selected  as  the consensus decision.

%As a probability distribution (which will be required later in our experiment to evaluate the loss values), the output from the majority vote rule will be given by 

In this work, we will compare the performance of fusion methods for consensus decision-making, assuming a symmetric loss function.
In this setting, the optimal decision for any fusion approach is to select the class with largest fused output.
In the case of ties, we will arbitrarily select the class with minimum index among those with maximal outputs.

\section{Classifier Fusion Method}
\label{chp:classification}
\label{sec:yayambo}
%The approach taken in this paper combines classifiers' outputs on an input vector $x_\textbf{in}$. %A classifier $f_j$ outputs $d_j$ on a given input vector $x_\text{in}$, which is denoted by $d_j=f_j(x_\text{in})$. For simplification in notation, $f_j(x_\text{in})$ is denoted as $d_j$.
%
%Assume there are $m$\label{classifier:m} classifiers, $f_i$, each embedded in a sensor node $r_i$\label{yayambo:ri}, and $l$ \label{classifier:l} class labels, $c_k$ \label{classifier:ck}. 
Given $x_\textbf{in}$, each classifier $f_i$'s output $d_i$\label{classifier:di} will serve as an initial distribution $\pi_i^{(0)}$\label{classifier:pi}.
These initial distributions are the input to the fusion algorithm, and they are updated over a number of iterations until consensus (or an iteration limit) is reached.
%As in~\cite{kittler1998combining}, $\pi_{ik}^{(0)}>0$, i.e. $\pi_{ik}^{(0)}\in (0, 1)$ for all $i$ and $k$.
%The intuition behind our approach is that since classifiers have learned to solve the same problem, there should be some dependency between classifiers' outputs.\snote{Removing - said before}
%To be able to determine reputability of classifiers, we assume some dependency between classifiers (i.e. between sensor nodes). 
%This dependency is motivated by the fact that classifiers have been constructed for the same task.\snote{Removed - redundant}
%We propose an iterative method for classifier fusion where a central node collects initial beliefs from some distributed sensor nodes and iteratively updates them until they converge.\snote{Removed - redundant}
%
We call the resulting fusion method the Yayambo\footnote{Yayambo means \textit{the first} in Kikongo, one of the languages spoken in the Democratic Republic of Congo. %I called this algorithm Yayambo because is the first research project with novelty I developed.
} algorithm.

During each iteration $t$, each current distribution $\pi_i^{(t-1)}$ is updated to a new distribution $\pi_i^{(t)}$\label{yayambo:belief}.
These updates are performed based on a vector of non-negative weights computed between each pair of current distributions.
Each vector component expresses a level of support for one distribution's prediction for a specific class from another, and is called a \textit{class support}.
Such a vector of class supports is termed a \textit{support}.
%omputes a vector of supports, called a class support vector, for each pair of distributions. We call a class support a weight assigned to a class. These class supports are collected into a class support vector, which we will refer by the term \textit{support}.   %$r_i$ evaluated from its belief and its neighbour $r_j$'s belief.  %sends a message $\langle \texttt{SupportMessage($\beta_{ij}^{(t)}$)}\rangle$ containing its  support $\beta_{ij}^{(t)}$\label{classifier:beta} to each neighbour $r_j$ and receives a message $\beta_{ji}^{(t)}$ from each neighbour $r_j$. %The sensor node $r_i$ then  considers each belief received from each of its neighbours to evaluate the support $\beta_{ji}^{(t)}$\label{classifier:beta} that the neighbour $r_j$ ought to confer to it. 
%These supports are 
%The prior belief of a sensor node is the output from its classifier (a prior belief is assumed to be with strictly positive probabilities), 
%\begin{equation}
%\pi_j^{(0)}=d_j\,.
%\end{equation}
%
%Let $\pi_i^{(t)}$ denote the $i$th distribution at iteration $t$. 
We denote the support \textit{for} $\pi_i^{(t-1)}$ \textit{from} $\pi_j^{(t-1)}$ by $\beta_{ji}^{(t)}$\label{classifier:beta}. The updated distribution $\pi_i^{(t)}$ is then calculated using the supports for $\pi_i^{(t-1)}$:
%The combiner uses these supports to update distributions $\pi_i^{(t-1)}$ to $\pi_i^{(t)}$ \label{yayambo:belief} as follows. 
%\begin{equation}
%\beta_{ij}^{(t)}=(\beta_{ij,1}^{(t)}, \beta_{ij,2}^{(t)}, \cdots, \beta_{ij,l}^{(t)})\,.
%\end{equation}
%The belief of sensor node $r_j$ at iteration $t$ is given as follows,
%\begin{equation}
%\label{eq:update}
%\pi_j^{(t)}=(\pi_{j1}^{(t)}, \pi_{j2}^{(t)}, \cdots, \pi_{jl}^{(t)})\,,
%\end{equation}
%and each $\pi_{jk}^{(t)}$ is updated in each iteration using the rule
\begin{equation}
\label{eq:update_yayambo}
\pi_{ik}^{(t)}=\alpha_i^{(t)} \pi_{ik}^{(t-1)}\sum_{j\neq i} \beta_{ji,k}^{(t)}\,,
\end{equation}
where the subscript $k$ denotes a vector component (one per class), and $\alpha_i^{(t)}$ is a normalisation constant.
Since the supports will be non-negative (as discussed later), this yields valid probability distributions at each iteration.
%When consensus is reached, the algorithm terminates: this 
Consensus occurs when these distributions stabilize: our implementation requires
%at the first iteration $t$ where $\pi_{i}^{(t)}\approx \pi_{i}^{(t-1)}$ for all $i$---% at iteration $t$, or at a predefined maximum iteration $T$.  
%our implementation uses the condition
\begin{equation}
\label{eq:error}
\sum_{i=1}^m\|\pi_{i}^{(t)}-\pi_{i}^{(t-1)}\|_2<m\varepsilon
\end{equation}
(for $\varepsilon=10^{-6}$).
%for a predefined small $\varepsilon=10^{-6}$.
On termination, we combine the final distributions using the sum rule---this simply provides a mechanism for resolving potential cases that do not reach consensus by the iteration limit.

%Since the algorithm might not reach consensus, two situations can arise. 
%When convergence occurs, say at iteration $T$, 
%the consensus belief is then,
%\begin{equation}
%\label{eq:consensus}
%\varpi \equiv \frac{1}{m} \sum_{j=1}^m d_j=\frac{1}{m}\bigg( \sum_{j=1}^m \pi_{i1}^{(T)}, \sum_{j=1}^m \pi_{i2}^{(T)}, \cdots, \sum_{j=1}^m \pi_{il}^{(T)}\bigg)\,.
%\varpi= \pi_{i}^{(T)}\,.
%\end{equation} 
%Note that, for the consistency of our method, we also use Equation \ref{eq:consensus} when convergence occurs which in any case still yields the same results, i.e. when all sensor nodes have the same final beliefs, the average of their beliefs (obtained from Equation \ref{eq:consensus}) will not change as compared to their final beliefs.
%where $\alpha$ denotes the normalisation constant.
%The description of the proposed method considers an ensemble of classifiers. If only one classifier is considered, the classifier just resorts to its own decision.

%Our proposed method, although not an instance of the loopy belief propagation algorithm \cite{murphy1999loopy}, is somewhat similar to it. Both methods have similarities in their convergence behaviour and update mechanisms. There is one major difference between the two methods. In our method, agents transfer support to each other while agents transfer beliefs about marginal distributions in the loopy belief propagation algorithm.

We finally discuss how the required supports $\beta_{ji}^{(t)}$ are calculated based on the current distributions $\pi_j^{(t-1)}$ and $\pi_i^{(t-1)}$.
%The final detail to complete the algorithm is specifying how to design support to score distributions. 
%
A support is a vector of positive scores that encodes a notion of similarity between two distributions: when two distributions are more similar, they assign stronger support to each other, as reflected by larger class supports.
%Support influences how much other distributions change. 
Support of distributions need not to be symmetric.
While various support functions could be defined, the function we consider
%\footnote{One may choose other dissimilarity measures. For instance, the Tversky measure is another dissimilarity measure which was originally designed to compare prototypes to variants \cite{tversky1977features}.}
is inspired by Kullback-Leibler divergence~\cite{kullback1951information}. %\cite{bregman1967relaxation, kullback1951information, murphy2012machine}.%, an entropy-based divergence. %Kullback-Leibler divergence is appropriate as a measure of dissimilarity between probability distributions and has also a likelihood interpretation: minimizing Kullback-Leibler divergence between two random variables is equivalent to maximizing the likelihood of the two random variables \cite{murphy2012machine}. 
 %The key motivation for this form of support is to be able to update a probability distribution based on the value of each of its component. In this way, the weights (or supports) applied to update distributions will be different (or non-uniform) across the components of a probability distribution. It will then be interesting that, after some iterations, distributions being updated with such non-uniform weights with respect to their components converge.
 
% This notion of support is related to the concept of Matthew Effect which, more abstractly, denotes that group members that have the same belief about something amplify their positions and collect ever-larger support from each other \cite{morton1968matthew}. In the perspective of our work, given a number of stakeholders (classifiers) divided into groups according to their relative similar distributions, members of each group will assign stronger support to each other, and assign weaker support to members of other groups.

This notion of support is related to the Matthew Effect~\cite{morton1968matthew}, where group members with similar beliefs about a topic amplify and collect ever-larger support from each other. In the setting of our work here, if we cluster stakeholders (classifiers) into groups whose predictions are similar, the members of each group will assign stronger support to each other, and assign weaker support to members of other groups.  %An important difference in our work, however, is that we consider similarity on each class independently, rather than on the prediction as a whole.

\begin{figure*}[h]
	\centering
	\begin{subfigure}[h]{0.3\textwidth}
		\includegraphics[width=\textwidth]{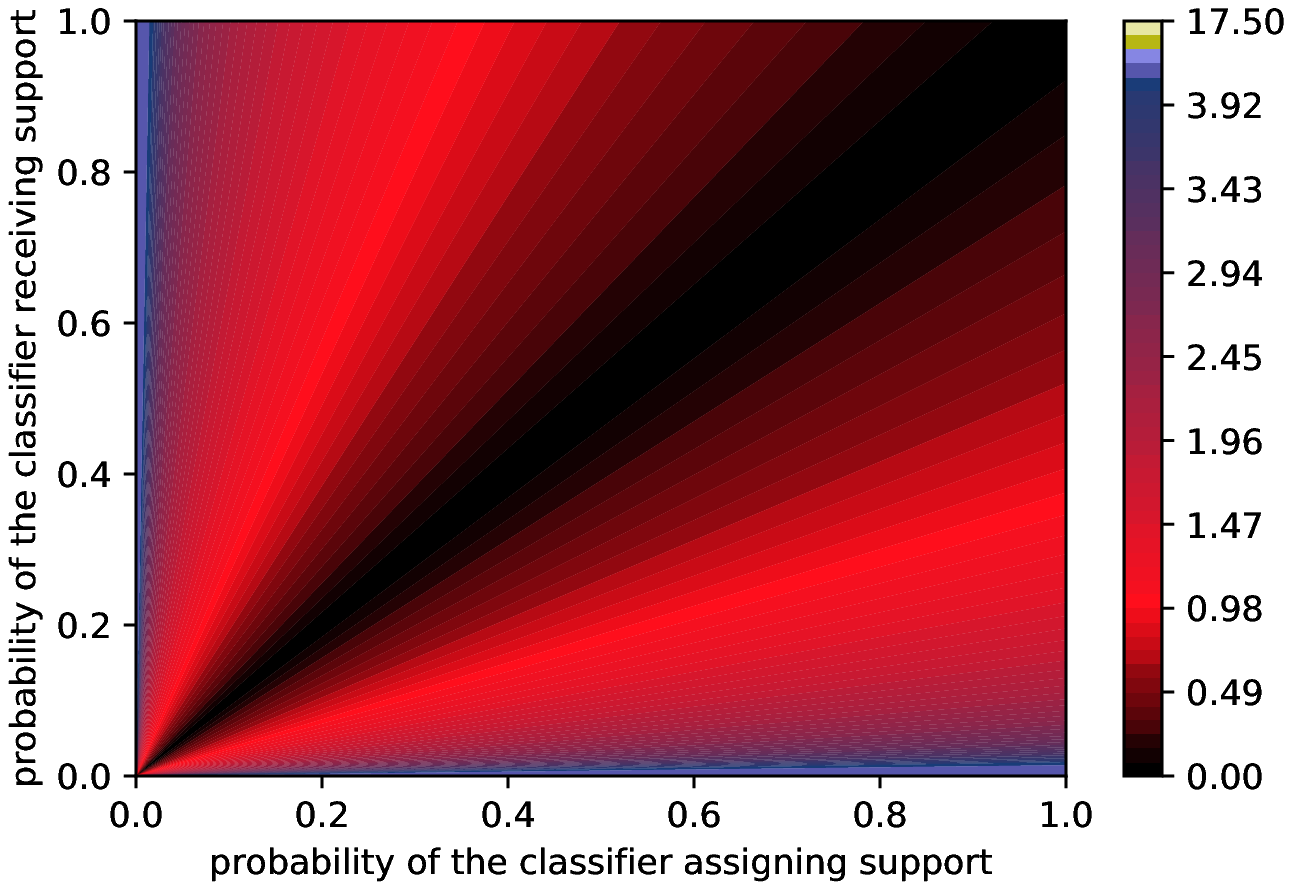}
		\caption{}
		\label{fig:similarityabs}
	\end{subfigure}
	\begin{subfigure}[h]{0.3\textwidth}
		\includegraphics[width=\textwidth]{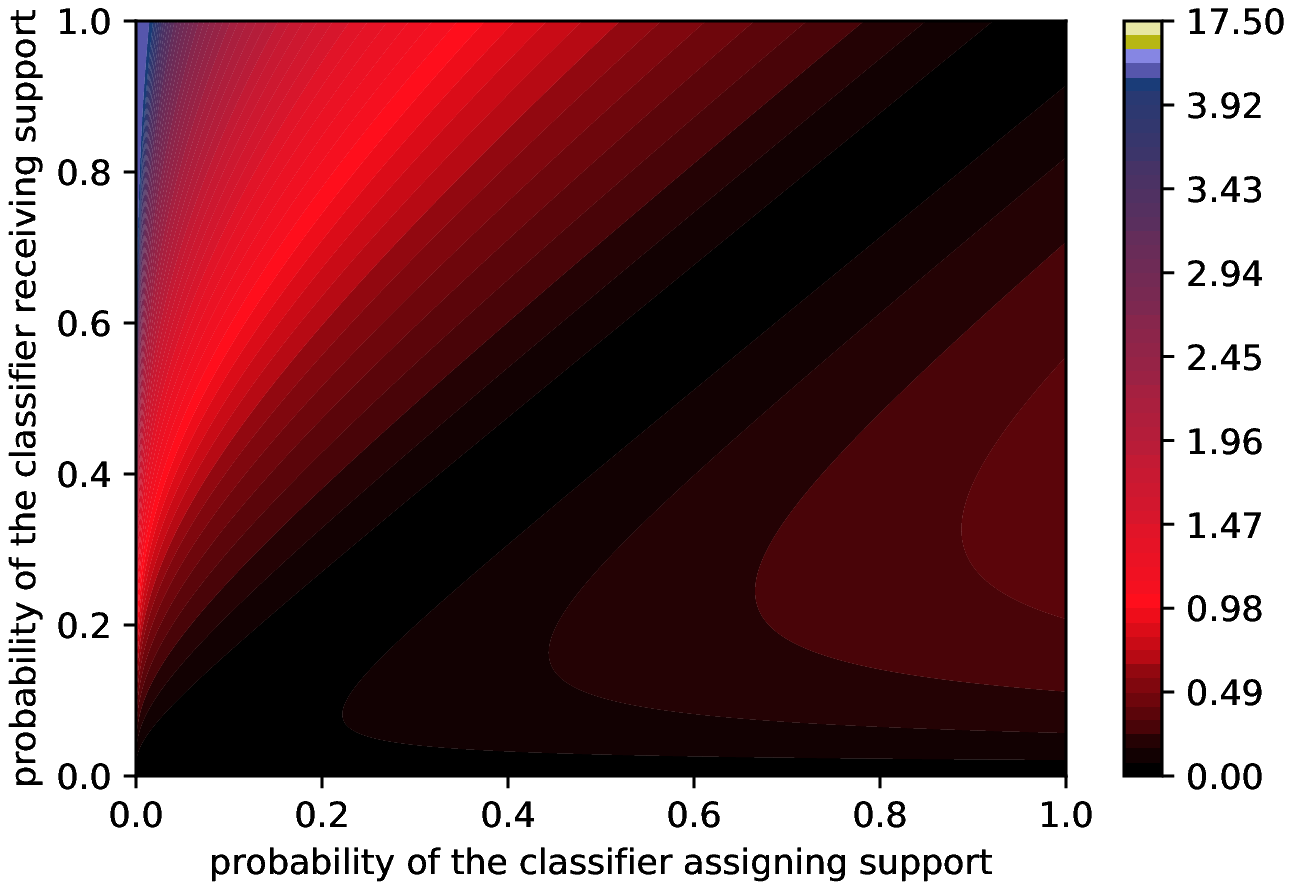}
		\caption{}
		\label{fig:similarityabs1}
	\end{subfigure}
	\begin{subfigure}[h]{0.3\textwidth}
		\includegraphics[width=\textwidth]{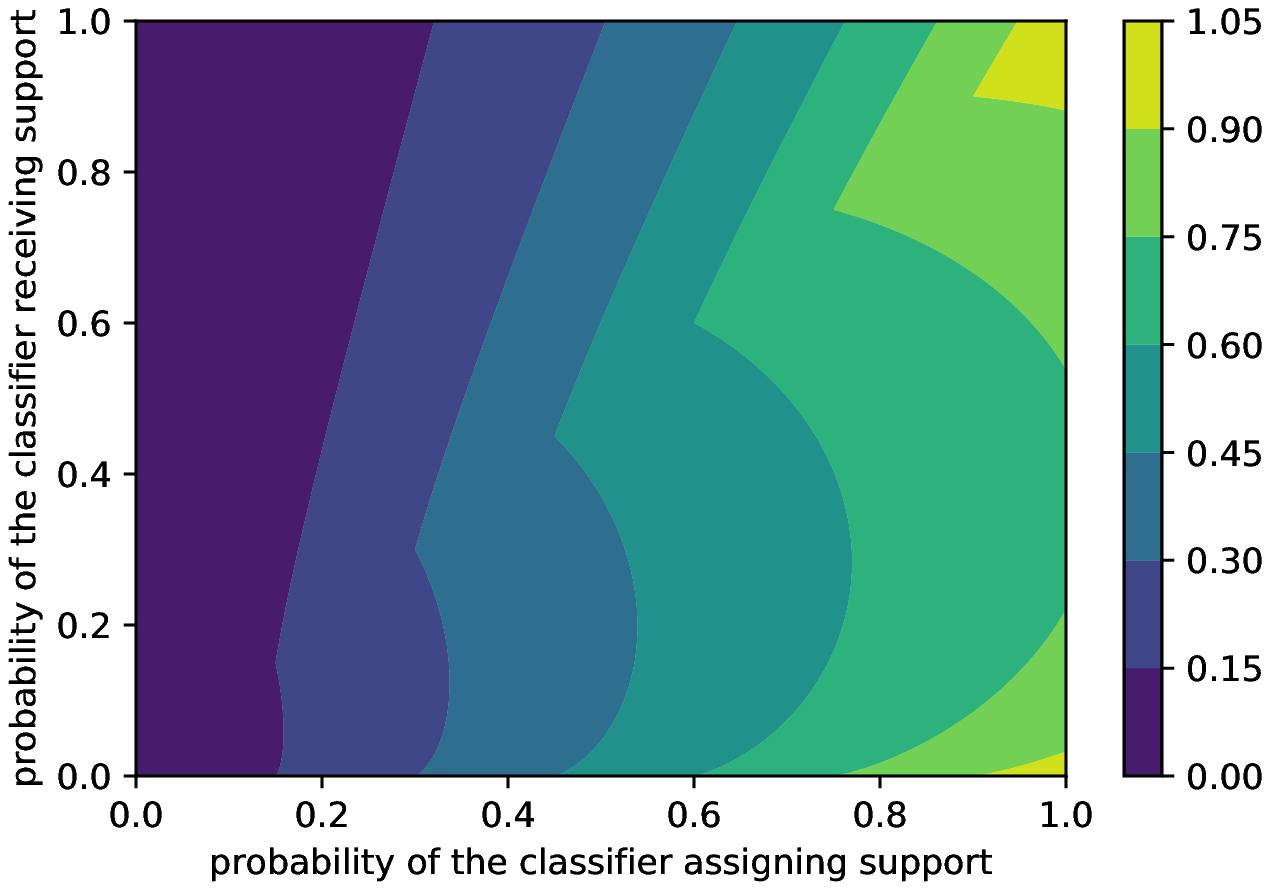}
		\caption{}
		\label{fig:similarities}
	\end{subfigure}
	\caption{Various quantities involved in calculation of class supports in Equation~\ref{eq:support}. \textbf{(\ref{fig:similarityabs})}: The absolute value of the log-ratio defined in the dissimilarity formula (Equation \ref{eq:divergence}). \textbf{(\ref{fig:similarityabs1})}: The dissimilarity defined in Equation~\ref{eq:divergence}.
				\textbf{(\ref{fig:similarities})}: Class support between probabilities as per Equation~\ref{eq:support}.}
	\label{fig:similarity}
\end{figure*}

The class support for $\pi_i^{(t-1)}$ from $\pi_j^{(t-1)}$ with respect to class $c_k$ depends principally on two things: first, the probability $\pi_j^{(t-1)}$ assigns to class $c_k$ (larger values corresponds to higher class supports), and second on the difference between the probabilities that $\pi_i^{(t-1)}$ and $\pi_j^{(t-1)}$ assign to class $c_k$ (larger differences lead to smaller class supports).
%Probability distributions are weighted in order to identify the most likely class of an object of interest. 
%In order to distinguish object types, a dissimilarity measure is applied to each component of each probability distribution.  
%The combiner wishes to work out which class is most likely to be the consensus class.
%The combiner strengthens distributions' supports on that class, while decreasing their supports on other classes.
Specifically, at iteration $t$, the class $k$ support for $\pi_i^{(t-1)}$ from $\pi_j^{(t-1)}$ is
\begin{equation}
\label{eq:support}
\beta^{(t)}_{ji, k}=\frac{\pi^{(t-1)}_{jk}}{1+D(\pi_{ik}^{(t-1)}, \pi_{jk}^{(t-1)})}\,,
\end{equation}
where the \textit{dissimilarity} for the $k$th components of $\pi_i^{(t)}$ and $\pi_j^{(t)}$ is given by
\begin{equation}
\label{eq:divergence}
D(\pi_{ik}^{(t)}, \pi_{jk}^{(t)})=\pi^{(t)}_{ik} |\log \frac{\pi^{(t)}_{ik}  +\varepsilon_0 }{\pi^{(t)}_{jk} +\varepsilon_0} |\,,
\end{equation}
and $\varepsilon_0=10^{-6}$ is a smoothing term to avoid zero division. Figure~\ref{fig:similarity} plots the class supports and related quantities in terms of the relevant class $k$ probabilities.

%In this case, each sensor node assigns itself high (in fact its own belief) self-support.
%\footnote{There are situations in which low self-support arises and must be handled. EasyChair is an example which a reviewer can assign low self-support \cite{voronkov2004easychair}.
%} %Since a support is a vector, it is high when all its values are high.
%\begin{algorithm}
%	\begin{algorithmic}[1]
%		\scriptsize
%		\STATE $\pi_i^{(t)}\gets (0, 0, \cdots, 0)$
%		\STATE $\alpha_i^{(t)} \gets 0$
%		\FORALL{sensor node $r_j\neq r_i$}
%		%\STATE the sensor node $r_i$ receives a message $\langle \texttt{SupportMessage($\beta_{ji}^{(t-1)}$)}\rangle$ from $r_j$
%		
%		\FORALL{$k=1$ to $l$}
%		\STATE $\pi_{ik}^{(t)}\gets \pi_{ik}^{(t)} + \pi_{ik}^{(t-1)}\beta^{(t)}_{ji, k}$
%		\STATE $\alpha_i^{(t)} \gets \alpha_i^{(t)} + \pi_{ik}^{(t)}$
%		\ENDFOR
%		\ENDFOR
%\STATE $\pi^{(t)}_i\gets \frac{\pi^{(t)}_i}{\alpha_i^{(t)}}$
%	\end{algorithmic}  
%	\caption{\scriptsize \normalsize{\texttt{update()}: \textit{This algorithm is a subroutine of Algorithm \ref{algo:yayambo} run by the central node $r$ to update the belief of sensor node $r_i$. Some variables used here are global, i.e are declared and initialised  in Algorithm \ref{algo:yayambo}.} }}
%	\label{algo:update_yayambo}
%\end{algorithm}
%
%The procedure that the central node applies to update the beliefs of sensor nodes is given in Algorithm \ref{algo:update_yayambo}. 
%We call this fusion method the Yayambo algorithm, and it

\begin{algorithm}
	%\hspace*{\algorithmicindent} \textbf{Input:} $ \varepsilon, T, \varepsilon_0, \pi^{(0)}_i \forall i$\\
	%\hspace*{\algorithmicindent} \textbf{Output:} $ \varpi$
	\begin{algorithmic}[1]
		\scriptsize
		\STATE $t\gets 0$
        \STATE $m\gets \textrm{number of distributions $\pi^{(0)}_i$ to fuse}$
		%\STATE $\pi^{(0)}_i\gets \pi^{(0)}_i$\label{line:yayambo_initialisation}
		\REPEAT\label{line:yayambo_loop_begin}
		\STATE $t\gets t + 1$
		\STATE $\texttt{difference}\gets0$
		\FORALL{$i$}
		\FORALL{$j\neq i$}
		\FORALL{$k=1$ to $l$}
		\STATE $\beta^{(t)}_{ji, k}\gets \frac{\pi^{(t-1)}_{jk}}{1+\pi^{(t-1)}_{ik} |\log \frac{\pi^{(t-1)}_{ik} +\varepsilon_0}{\pi^{(t-1)}_{jk}+\varepsilon_0}|}$
		\ENDFOR
		
		%\STATE the sensor node $r_i$ sends a message $\langle \texttt{SupportMessage($\beta_{ji}^{(t)}$)}\rangle$ to $r_j$
		\ENDFOR
		\STATE $\pi_i^{(t)}\gets (0, 0, \cdots, 0)$
		\STATE $\alpha_i^{(t)} \gets 0$
		\FORALL{$j\neq i$}
		%\STATE the sensor node $r_i$ receives a message $\langle \texttt{SupportMessage($\beta_{ji}^{(t-1)}$)}\rangle$ from $r_j$
		
		\FORALL{$k=1$ to $l$}
		\STATE $\pi_{ik}^{(t)}\gets \pi_{ik}^{(t)} + \pi_{ik}^{(t-1)}\beta^{(t)}_{ji, k}$
		\STATE $\alpha_i^{(t)} \gets \alpha_i^{(t)} + \pi_{ik}^{(t)}$
		\ENDFOR
		\ENDFOR
		\STATE $\pi^{(t)}_i\gets \frac{\pi^{(t)}_i}{\alpha_i^{(t)}}$
		\STATE $\texttt{difference}\gets\texttt{difference}+ \|\pi_{i}^{(t)}-\pi_{i}^{(t-1)}\|_2$
		\ENDFOR
		\UNTIL {$\texttt{difference}< m\varepsilon$ \OR $t==T$}\label{line:yayambo_end}\label{line:yayambo_loop_end}
		
		\FORALL{$k=1$ to $l$}
		\STATE $\varpi_k\gets 0$
		\FORALL{$i$}
		\STATE $\varpi_k\gets \varpi_k + \pi_{ik}^{(t)}$
		\ENDFOR
		\STATE $\varpi_k\gets \frac{\varpi_k}{m}$
		\ENDFOR
		%\STATE $\pi_\text{post}\gets \pi_{i}^{(t)}$
		%		\FORALL{neighbour $r_j$ of $r_i$}
		%		\STATE $\pi_\text{post}\gets \pi_\text{post} + \pi_{j}^{(t-1)}$
		%		\ENDFOR
		%		\STATE $\pi_\text{post}\gets \frac{1}{m} \pi_\text{post}$
		\RETURN $\varpi$
	\end{algorithmic}  
	\caption{\scriptsize \normalsize{\texttt{Yayambo()}}\textit{: The algorithm receives as inputs the classifier outputs $\pi_i^{(0)}$, a small threshold $\varepsilon_0$ to avoid zero division, the convergence threshold $(\varepsilon)$ and the maximum number of iterations $(T)$, and outputs the fusion result $\varpi$}.}
	\label{algo:yayambo}
\end{algorithm}
The resulting algorithm is detailed in Algorithm~\ref{algo:yayambo}.  The source code of Yayambo can be found at \url{https://bitbucket.org/jmf-mas/codes/src/master/classifier}. 
Assuming that $T$ is the number of iterations until termination, its complexity is $\mathcal{O}(T m^2l)$.  In terms of computational cost, the Yayambo method is more costly
than the benchmark fusion methods. However, for typical $m$ and $T$, this complexity is hardly
noticeable.
%To illustrate the functioning of the algorithm, let $f_1$ and $f_2$ be two binary classifiers with initial outputs $$\pi_1^{(0)}=(0.3, 0.7)\;\;\text{ and }\;\; \pi_2^{(0)}=(0.8, 0.2)\,,$$	where we set the convergence threshold $\varepsilon$ to $10^{-6}$. 
%It was mentioned that, to use our proposed method, it should be assumed that sensor nodes start with some disagreement in their initial beliefs, and we use an asymmetric measure to exhibit such disagreement\textemdash there is disagreement between sensor nodes when their belief on the same class differ (this disagreement would not be expressed if a symmetric measure was used to computer support). 
%From the initial beliefs of $r_1$ and $r_2$ above illustrated, it can be seen that the two sensor nodes disagree with each other on the class of the object of interest: $r_1$ has high belief for the second class while $r_2$ assigns high belief to the first class. 
Table~\ref{tab:yayambo_beliefs} illustrates the algorithm, showing the evolution of two distributions $\pi_1^{(t)}$ and $\pi_2^{(t)}$ during execution.
In this example, convergence is reached after 7 iterations.%Assignment of supports stops at iteration $t=6$ because sensor nodes' beliefs have reached consensus at iteration $t=7$.
%\stodo{Section on convergence commented out, since removed from the thesis.}
	
	\begin{table*}[ht]
		\centering
		\scalebox{0.8}{
			\begin{tabular}{|c|c|c|c|c|}
				\hline
				\textbf{$t$} & $\pi_1^{(t)}$ & $\pi_2^{(t)}$ & $\pi^{(t)}$ & $\sum_{i=1}^2\|\pi_i^{(t)}-\pi_i^{(t-1)}\|$ \\
				\hline
				$0$ &$(0.3, 0.7)$&$(0.8, 0.2)$&$(0.55, 0.45)$&$-$\\
				\hline
				$1$ &$(0.7130, 0.2870)$&$(0.5458, 0.4542)$&$(0.6294, 03706)$&$0.9435$\\
				\hline
				$2$ &$(0.7394, 0.2606)$&$(0.7589, 0.2411)$&$(0.7491, 0.2509)$&$0.3387$\\
				\hline
				$3$ &$(0.8994, 0.1006)$&$(0.8992, 0.1008)$&$(0.8993, 0.1007)$&$0.4247$\\
				\hline
				$4$ &$(0.9876, 0.0124)$&$(0.9876, 0.0124)$&$(0.9876, 0.0124)$&$0.2498$\\
				\hline
				$5$ &$(9.9984 \times 10^{-1}, 1.5727\times 10^{-4})$&$(9.9984\times 10^{-1}, 1.5727\times 10^{-4})$&$(9.9984\times 10^{-1}, 1.5727\times 10^{-4})$&$0.0346$\\
				\hline
				$6$ &$(9.9999\times 10^{-1}, 2.4740\, 10^{-8})$&$(9.9999\times 10^{-1}, 2.4740\times 10^{-8})$&$(9.9999\times 10^{-1}, 2.4740\times 10^{-8})$&$0.0004$\\
				\hline
				$7$ &$(9.9999\times 10^{-1}, 6.1208\times 10^{-16})$&$(9.9999\times 10^{-1}, 6.1208\times 10^{-16})$&$(9.9999\times 10^{-1}, 6.1208\times 10^{-16})$&$6.9976\times 10^{-16}$\\
				\hline
		\end{tabular}}
		\caption{Evolution of probability distributions over iterations of the Yayambo algorithm.  The column labelled $\pi^{(t)}$ shows the result of the sum rule applied to the distributions at the end of each iteration, and the final column shows the quantity monitored for detecting convergence.
		}
		\label{tab:yayambo_beliefs}
	\end{table*}

\section{Experimental Investigation}
\label{chp:classification_results}
\subsection{Methodology}
We compare our proposed classifier fusion method to the Borda count, majority vote, product and sum rules, discussed in Section \ref{chp:literature_classification}.
%as benchmarks.
We consider five performance metrics, namely: accuracy, cross-entropy loss, precision, recall, and F$_1$-score.\footnote{Since we mostly consider multi-class tasks, we report average precision, recall, and F$_1$-score when considering each class as the true class.}
Since cross-entropy loss heavily penalizes confident misclassifications, we expect our approach may perform poorly relative to the benchmarks on this metric, since it generates highly confident predictions by design: unless performance is perfect, the incorrect predictions will lead to large loss values.
%Due to its probabilistic formulation, we expect the sum rule to outperform our proposed method in terms of cross-entropy loss, which corresponds to a negative log-likelihood loss for a probabilistic model.
However, our contention is that our method may outperform the other techniques in terms of accuracy.
Our experiments were performed using six data sets from the UCI Repository~\cite{asuncion2007uci}, and one from from the Columbia object image library~\cite{nene1996columbia}: Iris; Gesture; Activity recognition; Handwritten digits; Satellite; Occupancy and Columbia.
Some characteristics of these data sets are given in Table~\ref{tab:data}. %Further descriptions of the various data sets can be found in Appendix \ref{app:dataset}.

Each object in the Columbia data set was captured from $72$ different views, at $5^\circ$ intervals around the object. In our experiments, we only consider $15$ types of objects.  We converted the RGB images to sequences of $16384$ grayscale values, and then retained $97$ principal components.\footnote{We built the PCA model using all of the training data.}
\begin{table}[!t]
	\renewcommand{\arraystretch}{1.3}
	\centering
	\scalebox{0.7}{
		\begin{tabular}{|c|c|c|c|c|c|}
			\hline
			\textbf{Data set} & \textbf{Training} & \textbf{Test} & \textbf{Attributes} & \textbf{Classes}&$n$\\
			\hline
			Iris &$105$ &$45$&$4$&$3$&$2$\\
			\hline
			Gesture&$785$ &$336$&$19$&$5$ &$12$ \\
			\hline
			Activity recognition &$4725$ &$2025$&$6$&$7$ &$4$ \\
			\hline
			Handwritten digits &$10000$ &$3000$&$784$&$10$&$25$ \\
			\hline
			Satellite &$4435$ &$2000$&$36$&$6$&$299$ \\
			\hline
			Occupancy &$8143$ &$2665$&$5$&$2$&$2$ \\
			\hline
			Columbia &$765$ &$315$&$97$&$15$&$54$ \\
			\hline
	\end{tabular}}
	\caption{The characteristics of training data sets and the number of features considered to train classifiers on each data set where $n$ indicates the number of features selected (this will be used for situations where classifiers are trained on different subsets of features of a data set).}
	\label{tab:data}
\end{table}
%For the Columbia data set, which consists of images of objects, each object was captured from different views ($72$ views, of $5^\circ$ around the object, in total). It is expected that the more views of the object we consider, the better performance we can achieve. In our experiments, we will consider some limited number of views of the objects and see how the number of views of objects will impact the performance of the classifier fusion methods.

%\subsection{Experimental setup}
%
%In experimental investigations%\footnote{The source codes for Yayambo model can be found here, \href{https://bitbucket.org/jmf-mas/phdproject/src/experimental/experiments/yayambo/}{https://bitbucket.org/jmf-mas/phdproject/src/experimental/experiments/yayambo/}.}
%
For our experiments, we use Python and \texttt{scikit-learn}~\cite{pedregosa2011scikit}. %\footnote{An open library for machine learning.}
We generated classifier outputs using five classifiers trained with various hyperparameters:
%Any classifier which outputs a probability distribution over classes can be used directly.
%The classifiers which were considered in these experiments were:
\begin{itemize}
\item logistic regression fit with \texttt{liblinear}~\cite{fan2008liblinear} ($f_1$),
\item $k$-nearest neighbours ($k$-NN) ($f_2$),
\item a support vector machine (SVM) with the sigmoid function kernel ($f_3$),
\item an SVM with the radial basis function (RBF) kernel ($f_4$), and
\item a multi-layer perceptron (MLP) trained with stochastic gradient descent ($f_5$).
\end{itemize}
%\footnote{\texttt{liblinear} is a library for linear support vector machines.  It is used for large-scale linear classification.} 
The hyperparameters we selected to train these classifiers are given in Tables~\ref{tab:hyper} and~\ref{tab:hyper_similar_classifiers}---for other hyperparameters, we used default values provided by \texttt{scikit-learn}.
These hyperparameters were deliberately selected  to result in highly similar as well as highly diverse classifiers.
%We observe classifiers' performances for different hyperparameters and consider those for which classifiers have different behaviours. %, which means classifiers with different opinions on some observations. 
Classifiers $f_1$ and $f_5$ output valid probability distributions by default; the predictions from the support vector machines ($f_3$ and $f_4$) were converted to probability distributions using the softmax operator~\cite{bishop2006pattern}, while
%classifiers' outputs are provided in forms of probabilities over classes, except $k$-nearest neighbours.
%With the scikit-learn library, by  the use of softmax function
for $k$-nearest neighbours ($f_2$) the probability distribution used was the empirical distribution over the neighbours' classes. %We aimed to investigate how various classifiers with different performances can be combined using a message-passing algorithm.
%Classifiers are all trained on the same data and features.
\begin{table}[h]
	% increase table row spacing, adjust to taste
	\centering
	% Some packages, such as MDW tools, offer better commands for making tables
	% than the plain LaTeX2e tabular which is used here.
	\scalebox{0.7}{
		\begin{tabular}{|l|c|c|c|c|c|c|c|}
			\hline
			Techniques&Iris&Gesture&Activity& Satellite&Digits& Occupancy& Columbia\\
			\hline
			$f_1$ &$C=2$ &$C=30$ &$C=0.2$&$C=0.002$ &$C=0.05$&$C=0.002$&$C=0.002$ \\
			\hline
			$f_2$ &$k=50$ &$k=15$ &$k=2000$&$k=500$ &$k=150$&$k=500$&$k=80$ \\
			\hline
			$f_3$ &$C=200$ &$C=80$ &$C=0.05$&$C=0.03$ &$C=0.8$&$C=0.005$ &$C=0.005$\\
			\hline
			$f_4$ &$C=300$ &$C=20$ &$C=0.03$&$C=0.015$ &$C=0.5$&$C=300$ &$C=300$ \\
			\hline
			\multirow{3}{*}{$f_5$} & $N=5$ & $N=5$&$N=5$& $N=5$&$N=5$&$N=5$ &$N=5$\\
			& $\lambda=0.2$ & $\lambda=0.2$ & $\lambda=0.2$ &$\lambda=0.2$ & $\lambda=0.2$& $\lambda=0.2$ & $\lambda=0.045$ \\
			& $T=200$ & $T=200$ & $T=300$ &$T=300$ & $T=500$& $T=500$& $T=500$ \\
			\hline
	\end{tabular}}
	\caption{Hyperparameters used to train classifiers. For $f_1, f_3$ and $f_4$,
%which are support vector machines, 
$C$ denotes the penalty parameter of the loss term. For $f_2$, $k$ denotes the number of neighbours. For $f_5$, $N$, $\lambda$ and $T$ denote the number of hidden layers, the learning rate and the number of epochs.}
	\label{tab:hyper}
\end{table}
\begin{table}[h]
	% increase table row spacing, adjust to taste
	\centering
	% Some packages, such as MDW tools, offer better commands for making tables
	% than the plain LaTeX2e tabular which is used here.
	\scalebox{0.7}{
		\begin{tabular}{|l|c|c|c|c|c|}
			\hline
			Techniques&$h_1$&$h_2$&$h_3$& $h_4$&$h_5$\\
			\hline
			\multirow{3}{*}{MLP} & $N=10$ & $N=3$&$N=3$& $N=3$&$N=2$\\
			& solver=sgd & solver=adam&solver=lbfgs& solver=adam&solver=sgd\\
			& $\lambda=0.045$ & $\lambda=0.5$ & $\lambda=0.5$ &$\lambda=0.1$ & $\lambda=0.045$ \\
			& $T=500$ & $T=100$ & $T=100$ &$T=500$ & $T=500$ \\
			
			& $\sigma$=relu & $\sigma$=relu & $\sigma$=tanh &$\sigma$=logistic & $\sigma$=sigmoid \\
			\hline
			\multirow{3}{*}{$k$-NN} &
			$k=30$ & $k=60$&$k=90$& $k=120$&$k=150$\\
			& w=uniform& w=distance&w=uniform& w=uniform&w=distance\\
			& solver=ball tree& solver=ball tree&solver=kd tree& solver=brute&solver=brute\\
			\hline
			\multirow{3}{*}{SVM} & $C=3$ & $C=0.000001$&$C=0.1$& $C=100$&$C=1$\\
			& kernel=rbf & kernel=linear&kernel=poly& kernel=sigmoid&kernel=rbf\\
			& tol$=0.001$ & tol$=0.01$ & tol$=0.1$ &tol$=0.001$ & tol$=0.001$ \\
			& degree$=3$ & degree$=4$ & degree$=5$ &degree$=10$ & degree$=10$ \\
			\hline
	\end{tabular}}
	\caption{Hyperparameters used to train similar classifiers on the Columbia data set.}
	\label{tab:hyper_similar_classifiers}
\end{table}

Our experiments considered the following cases for each data set, in an attempt to comprehensively test the proposed approach and identify its limitations. (Unless otherwise specified, hyperparameters were as in Table~\ref{tab:hyper}):
\begin{itemize}
\item Fusing the outputs of $f_1$ to $f_5$ trained on the default training data sets.
\item Fusing the outputs of multiple similar classifiers $h_1$ to $h_5$ (either all MLPs, all $k$-NN, or all SVM classifiers) trained on the default training sets, with hyperparameters as per Table~\ref{tab:hyper_similar_classifiers}.
\item Fusing the outputs of $f_1$ to $f_5$ trained on differing subsets of observations from the default training sets.
\item Fusing the outputs of $f_1$ to $f_5$ trained on differing sets of features of the default training sets.
\item Fusing the outputs of varying numbers of classifiers (between 2 and 5) by selecting subsets of $f_1$ to $f_5$.
\end{itemize}
%The convergence threshold $\varepsilon$ in Equation~\ref{eq:error} was set to $10^{-6}$ throughout.  
Finally, we consider some artificial settings where our assumptions of regularity of trained classifiers may be violated.

%As was mentioned earlier that different orderings can result different performance, we evaluate all the possibilities (i.e. $m!=5!=120$ orderings for this work) and choose a best ordering. %The choice of a correct ordering of classifiers is an offline process, as it happens in batch. It does not affect the runtime of the model. 

%\begin{figure}
%	\centering
%	\includegraphics[scale=0.5]{figs/errorbars}
%	\caption[\textit{Error bars using different classifiers on each dataset.}]{\textit{Error bars using different classifiers on each dataset.  They represent the error in classifiers' accuracy. This gives a general idea of how good classifiers are, or conversely, how far from the correct decisions classifiers' decisions might be. We used the standard deviation to represent classifier error. }}
%	\label{fig:errorbars}
%\end{figure}
\subsection{Results and discussion}
\label{sec:fusion_results}
Over all the experiments, Yayambo reached consensus within 5--23 iterations.

\subsubsection{Classifiers trained on the same data set}
%
%individual performance
%Individual classifier performance is presented in Figure \ref{fig:errorbars}. 
%
The results of the five metrics we considered when applying the Borda count rule, the majority vote rule, the sum rule, the product rule and the Yayambo fusion technique to classifiers trained on the same data set are presented in Tables~\ref{tab:camaccuracy}--\ref{tab:camloss}.
%The means and standard deviations of classifiers' accuracies on data set are $0.9640\pm 0.0296$ for the Iris data set, $0.6583\pm0.0272$ for the Gesture data set, $0.6342\pm 0.0806$ for the Activity recognition data set, $0.6710\pm 0.0733$ for the Handwritten digits data set, $0.8839\pm 0.0336$ for the Satellite data set, $0.7453\pm 0.1352$ for the Occupancy data set and  $0.7232\pm 0.1320$ for the Columbia data set. 

While almost all the fusion algorithms exhibited perfect accuracy, precision, recall and F$_1$ values on the Iris test data set, the decisions resulting from the Yayambo method outperformed those from the benchmark methods on all the other data sets for all of these metrics.\footnote{The only exception was a tie in the precision of the product rule and Yayambo on the Activity data set.} Our results also confirm those of~\cite{kittler1998combining1} in that the sum rule generally outperformed the majority vote rule and the product rule on these metrics.
We see that all five fusion methods behave similarly when all of the individual classifiers are strong.
%In these cases, Yayambo behaves similarly to the sum rule and product rule methods.
%In Table~\ref{tab:camaccuracy} for instance, we observed that classifiers give almost the same opinions on data points in the Iris. The three fusion methods have very similar accuracy.
The Yayambo method is most beneficial when the predictions are highly diverse, i.e. the classifiers give different opinions for the same input.
This reflects in better fusion performance when combining classifiers  with more widely differing performance levels, %But it can also be applied where classifiers give the same decision for a given input. 
%A difference in performance is observed as expected when classifiers give more widely varying predictions, 
such as in the Columbia data set in Table \ref{tab:camaccuracy}.
%When classifiers were inconsistent, Yayambo outperformed the sum rule and product rule methods more notably in terms of accuracy. % given that agents' supports can converge after a number of iterations. 
%
% The sum rule, product rule and Yayambo methods give almost the same accuracy when classifiers yield almost the same performances. But Yayambo is the best (in terms of accuracy) when individual classifiers achieve different accuracies.
These results indicate that it is beneficial to use Yayambo as it provides robustness to varying quality of individual classifiers on various quality metrics, with comparable performance when classifiers have similar performance.

%Although Table \ref{tab:camaccuracy} also shows that SUM and Yayambo outperform individual classifiers, this need not always be the case.
%
\begin{table}[h]
	% increase table row spacing, adjust to taste
	\centering
	% Some packages, such as MDW tools, offer better commands for making tables
	% than the plain LaTeX2e tabular which is used here.
	\scalebox{0.7}{
		\begin{tabular}{|l|c|c|c|c|c|c|c|}
			\hline
			Techniques&Iris&Gesture&Activity& Satellite&Digits&Occupancy&Columbia\\
			\hline
			$f_1$ &$0.9111$ &$0.6845$ &$0.7491$&$0.5675$ &$0.9063$&$0.6353$&$0.8762$ \\
			$f_2$ &$0.9556$ &$0.6667$ &$0.6642$&$0.7885$ &$0.8753$&$0.9295$&$0.6635$ \\
			$f_3$ &$0.9778$ &$0.6488$ &$0.6104$&$0.6925$ &$0.9047$&$0.6353$&$0.5270$ \\
			$f_4$ &$0.9778$ &$0.6101$ &$0.6459$&$0.6785$ &$0.9117$&$0.8912$&$0.6444$ \\
			$f_5$ &$0.9778$ &$0.6815$ &$0.5012$&$0.6280$ &$0.8217$&$0.6353$&$0.7268$ \\
			\hline
			Borda count rule&$0.9777$ &$0.6455$ &$0.7356$&$0.8090$ &$0.8905$ &$0.9033$ &$0.8878$  \\
			Majority vote rule&$\textbf{1.0000}$ &$0.6728$ &$0.7446$&$0.8122$ &$0.8886$ &$0.9133$ &$0.8996$  \\
			Product rule &$\textbf{1.0000}$ &$0.6665$ &$0.7455$&$0.8110$ &$0.8905$ &$0.9042$ &$0.8852$  \\
			Sum rule &$\textbf{1.0000}$ &$0.6875$ &$0.7644$&$0.8010$ &$0.9153$ &$0.9163$ &$0.9048$  \\
			Yayambo &$\textbf{1.0000}$ &$\textbf{0.7054}$ &$\textbf{0.7788}$&$\textbf{0.8120}$ &$\textbf{0.9167}$&$\textbf{0.9365}$&$\textbf{0.9535}$ \\
			\hline
	\end{tabular}}
	\caption[]{Test accuracy of individual classifiers, the benchmark fusion methods and Yayambo for various data sets. %Accuracy of individual classifier and fusion techniques on benchmark data sets. %Yayambo outperforms any individual classifier in this case. 
			Bold values in all tables indicate the best performance.}
	\label{tab:camaccuracy}
\end{table}

\begin{table}[h]
	% increase table row spacing, adjust to taste
	\centering
	% Some packages, such as MDW tools, offer better commands for making tables
	% than the plain LaTeX2e tabular which is used here.
	\scalebox{0.7}{
		\begin{tabular}{ |l|l|l|l|l| }
			\hline
			Data set&Methods&Precision&Recall& F$_1$-score\\
			\hline
			\multirow{2}{*}{Iris} & Borda count rule& $0.9800$&$0.9800$&$0.9800$\\
			& Majority vote rule& $\bm{1.0000}$&$\bm{1.0000}$&$\bm{1.0000}$\\
			& Sum rule & $\bm{1.0000}$&$\bm{1.0000}$&$\bm{1.0000}$\\
			& Product rule & $\bm{1.0000}$&$\bm{1.0000}$&$\bm{1.0000}$\\
			& Yayambo & $\bm{1.0000}$&$\bm{1.0000}$&$\bm{1.0000}$ \\
			\hline
			\multirow{2}{*}{Gesture} & Borda count rule & $0.6700$ & $0.6800$ &$0.6100$\\
			& Majority vote rule & $0.6600$ & $0.6800$ &$0.6000$\\
			& Sum rule & $0.7000$ & $0.6900$ &$0.6500$\\
			& Product rule & $0.6900$ & $0.6900$ &$0.6400$\\
			& Yayambo & $\bm{0.7100}$&$\bm{0.7100}$&$\bm{0.6900}$ \\
			\hline
			\multirow{2}{*}{Activity} & Borda count rule & $0.7700$ & $0.7500$ &$0.7500$\\
			 & Majority vote rule & $\bm{0.7800}$&$\bm{0.7800}$&$\bm{0.7800}$\\
			& Sum rule & $0.7700$ & $0.7600$ &$0.7600$\\
			& Product rule & $\bm{0.7800}$&$0.7700$&$0.7600$ \\
			& Yayambo & $\bm{0.7800}$&$\bm{0.7800}$&$\bm{0.7800}$ \\
			\hline
			\multirow{2}{*}{Satellite} &Borda count rule & $0.7900$&$0.7900$& $0.7800$\\
			& Majority vote rule & $0.7800$&$0.7900$& $0.7900$\\
			& Sum rule & $0.8000$&$0.8000$& $0.8000$\\
			& Product rule & $0.7800$&$0.7800$& $0.7900$\\
			& Yayambo & $\bm{0.8200}$ & $\bm{0.8100}$ &$\bm{0.8100}$\\
			\hline
			\multirow{2}{*}{Digits} & Borda count rule & $0.9100$&$0.9100$& $0.9100$\\
			& Majority vote rule & $0.9100$&$0.9100$& $0.9100$\\
			& Sum rule & $0.9100$&$0.9100$& $0.9100$\\
			& Product rule & $0.9100$&$0.9100$& $0.9100$\\
			& Yayambo & $\bm{0.9200}$ & $\bm{0.9200}$ &$\bm{0.9200}$\\
			\hline
			\multirow{2}{*}{Occupancy} & Borda count rule & $0.9000$&$0.9100$& $0.9200$\\
			& Majority vote rule & $0.9000$&$0.9000$& $0.9200$\\
			& Sum rule & $0.9300$&$0.9200$& $0.9200$\\
			& Product rule & $0.9000$&$0.9100$& $0.9200$\\
			& Yayambo & $\bm{0.9600}$ & $\bm{0.9500}$ &$\bm{0.9500}$\\
			\hline
			\multirow{2}{*}{Columbia} & Borda count rule& $0.9200$&$0.9000$& $0.8900$\\
			& Majority vote rule& $0.9300$&$0.8900$& $0.8800$\\
			& Sum rule & $0.9300$&$0.9000$& $0.9000$\\
			& Product rule & $0.9300$&$0.8900$& $0.8800$\\
			& Yayambo & $\bm{0.9500}$ & $\bm{0.9400}$ &$\bm{0.9400}$\\
			\hline
	\end{tabular}}
	%\caption[\textit{Other performance metrics on classifier fusion algorithms.}]{\textit{Precision, recall (the true positive rate), and F$_1$ score for the fusion techniques on the seven benchmark data set. The Yayambo approach performed best on all metrics for all these data sets. Bold values indicate best performance.}}
	\caption[]{Precision, recall, and F$_1$ values of individual classifiers, the benchmark fusion methods  and Yayambo for various data sets. %Accuracy of individual classifier and fusion techniques on benchmark data sets. %Yayambo outperforms any individual classifier in this case. 
		%	Bold values indicate the best performance.
		}
	\label{tab:othermetric}
\end{table}

As expected, the sum rule and product rule outperformed Yayambo in terms of cross-entropy loss in all cases except the Iris data set, where no prediction errors were made.
%When considering cross-entropy loss, the sum rule and product rule outperform Yayambo. 
This is because the sum rule and product rule methods each return a sort of average of the probability distributions of individual classifiers, so that these two methods will not return overconfident predictions.
%In such cases, the loss is limited because 
Thus, in the case of misclassified observations, the contribution to the loss with the sum rule and product rule are limited.
On the other hand, Yayambo's consensus-seeking approach leads to overconfident classifications, with extremely high corresponding loss values in the case of misclassification: this results from the probability of the correct class being driven down to zero.
These high loss values typically easily outweigh the reduction in loss caused by more confident correct classifications. %This result of having high loss values also applies to  Majority vote rule 1 because this majority vote rule does not output a calibrated  distribution.
%  reports a high loss on misclassified points.
%This is because Yayambo does not average the classifiers' probability distribution, but assigns a score of close to $1$ to a consensus class label and scores of almost zero to other class labels. Thus, when an observation is misclassified, Yayambo returns a large loss. %The only time Yayambo can ensure loss-free performance is when it has $100\%$ accuracy\textemdash as observed in the case of the Iris data set in Table \ref{tab:camloss}.
%This indicates that it is not recommended to use Yayambo in situations which require fusion methods with well-calibrated outputs.
This  behaviour confirms that our fusion approach is not suitable for downstream tasks where calibrated fused output probabilities are desired.
It is also worth noting that for most data sets, one or more of the individual classifiers had lower cross-entropy loss than all the fusion methods.

\begin{table}[h]
	% increase table row spacing, adjust to taste
	\centering
	% Some packages, such as MDW tools, offer better commands for making tables
	% than the plain LaTeX2e tabular which is used here.
	\scalebox{0.7}{
		\begin{tabular}{|l|c|c|c|c|c|c|c|}
			\hline
			Model&Iris&Gesture&Activity& Satellite&Digits&Occupancy&Columbia\\
			\hline
			$f_1$ &$0.4798$ &$0.8798$ &$0.9229$&$1.3969$ &$0.4225$&$0.5783$ &$2.7037$ \\
			$f_2$ &$0.4451$ &$1.1447$ &$1.2374$&$\textbf{0.5073}$ &$0.4451$&$\textbf{0.1597}$&$\textbf{0.8634}$ \\
			$f_3$ &$0.0891$ &$0.8817$ &$0.7077$&$0.5501$ &$0.2899$&$0.2133$&$2.7086$ \\
			$f_4$ &$0.1125$ &$0.8791$ &$\textbf{0.6057}$&$0.5532$ &$\textbf{0.2720}$&$0.3716$&$1.2267$ \\
			$f_5$ &$0.0495$ &$0.8786$ &$0.6889$&$0.8709$ &$0.7405$&$0.7304$&$6.7406$ \\
			\hline
			Borda count rule &$0.0512$ &$0.8574$ &$1.2443$&$1.4677$ &$1.0010$ &$0.9000$ &$1.2442$  \\
			Majority vote rule &$0.2012$ &$0.8657$ &$0.8888$&$1.2282$ &$0.6767$ &$1.1989$ &$1.2111$  \\
			Product rule &$0.1021$ &$0.8541$ &$0.9542$&$1.2564$ &$0.9985$ &$1.2564$ &$1.2020$  \\
			Sum rule &$0.2135$ &$\textbf{0.7485}$ &$0.7586$&$0.6459$ &$0.3304$ &$0.2873$ &$1.1151$  \\
			Yayambo &$\textbf{0.0000}$ &$10.7262$ &$3.5044$&$3.0495$ &$1.6125$&$0.6944$&$1.9882$ \\
			\hline
	\end{tabular}}
	\caption[]{Cross-entropy losses of individual classifiers, the benchmark fusion methods and Yayambo for various data sets. %Accuracy of individual classifier and fusion techniques on benchmark data sets. %Yayambo outperforms any individual classifier in this case. 
		%	Bold values indicate the best performance.
		}
	\label{tab:camloss}
\end{table}

%But sometimes, it is not possible to maximize all these performance metrics at the same time, as one comes at the cost of another\textemdash trade-off. For problems where there is a trade-off between some of these performance metrics, a decision has to be made whether to maximize which performance metric. For example if there is a trade-off between recall and precision, a decision could consist of using $F_1$ score, which is a harmonic mean of precision and recall.
The Borda count rule and majority vote rule were outperformed by other fusion methods in various experiments run. A potential weakness of the Borda count rule and majority vote rule is that the specific values of the probabilities are ignored when performing the fusion (although this might make them robust to outliers or overconfident classifiers).

In Tables \ref{tab:camaccuracy}--\ref{tab:camloss}, we considered five fusion methods.
Since in these tests the sum rule generally outperforms the product rule, the Borda count rule and the majority vote rule, the results in the following sections will focus almost exclusively on Yayambo and the sum rule.
%In what follows, we will only present results for Yayambo and the sum rule. We want to focus on Yayambo and the sum rule because the sum rule was reported to be the best fusion method \cite{kittler1998combining} for methods based on principle of indifference\textemdash it was also found in our experiments that the product rule typically achieves poorer results than the sum rule and Yayambo. 
Similarly, in our further experiments, we will only report on accuracy: we have already established that Yayambo is vulnerable to poor performance on cross-entropy loss.
%In our experiments, Yayambo was found to achieve the best on the other three metrics. So to avoid repeating the same conclusions, the results on these other metrics will be omitted.

\subsubsection{Similar classifiers trained with different parameters on the same data set}
Results of the fusion methods on the Columbia data set using the same classifiers but with different hyperparameters are presented in Table~\ref{tab:camaccuracy_similar}.
Here $h_1$ to $h_5$ are all classifiers of the same type---denoted by the row heading---with hyperparameters as specified in Table~ \ref{tab:hyper_similar_classifiers}.
%Table \ref{tab:hyper_similar_classifiers} shows the hyperparameters used when training the same classifier but with different hyperparameters.
%For this we consider three different techniques (MLP, $k$-NN and SVM). %Yayambo  achieved the best fused accuracy for each classifier type.
\begin{table}[h]
	% increase table row spacing, adjust to taste
	\centering
	% Some packages, such as MDW tools, offer better commands for making tables
	% than the plain LaTeX2e tabular which is used here.
	\scalebox{0.7}{
		\begin{tabular}{|l|c|c|c|}
			\hline
			Techniques&MLP&$k$-NN&SVM\\
			\hline
			$h_1$ &$0.6444$ &$\textbf{0.8381}$ &$0.6444$ \\
			$h_2$ &$0.4635$ &$0.8127$ &$0.7937$ \\
			$h_3$ &$0.4667$ &$0.5937$ &$0.9534$ \\
			$h_4$ &$0.4857$ &$0.4762$ &$0.6095$ \\
			$h_5$ &$0.1238$ &$0.7400$ &$0.1048$ \\
			\hline
			Borda count rule &$0.6052$ &$0.6873$ &$0.9016$  \\
			Majority vote rule &$0.7014$ &$0.7714$ &$0.7968$ \\
			Product rule &$0.6557$ &$0.7714$ &$0.9778$  \\
			Sum rule &$0.7486$ &$0.7841$ &$0.9810$  \\
			Yayambo &$\textbf{0.8000}$ &$0.7841$ &$\textbf{0.9905}$ \\
			\hline
	\end{tabular}}
\caption{Test accuracy of the fusion methods and individual classifiers using similar classifiers with different hyperparameters trained on the Columbia data set.
	%{\textit{The sum rule and Yayambo accuracies on test data using similar classifiers with different hyperparameters trained, on the Columbia data set where $h_i$ denotes the $i$-th instance of the corresponding classification method. Accuracy of individual classifier and fusion techniques on the Columbia data set. %Yayambo outperforms any individual classifier in this case. 
		%	Bold values indicate best performance.
		 }
	\label{tab:camaccuracy_similar}
\end{table}
%Table \ref{tab:camaccuracy_similar} shows that, when we fuse outputs from the same classifier but with different hyperparameters, Yayambo yields the best accuracy in general. 
The fusion methods yielded the same accuracy when fusing multiple $k$-NN classifiers,  with two of the individual classifiers outperforming all the fusion approaches.
Yayambo outperformed the sum rule for fusing MLP and SVM classifiers, where both fusion methods outperformed all the individual classifiers and the other three fusion methods.
%In Table \ref{tab:camaccuracy_similar}, individual classifiers have widely differing outputs for the test observations and generally have poor performance. The fusion methods are able to ameliorate  this inconsistency of classifiers to achieve better prediction. \stodo{Removed this, since it is not comparing the fusion approaches.}

%Two situations could be considered to explain the (relative) performances of Yayambo and the sum rule based on differences observed in classifiers' performances.\stodo{Removing this paragraph because it is repeating earlier discussions, essentially.}
%If the difference observed in classifiers' accuracies does not mean that classifiers provide different views on some data points, i.e. classifiers are consistent, it should be expected that good classifier fusion methods outperform individual classifiers.\stodo{Why should we expect this?}
%On the other hand, if differences observed in classifiers' performances indicate stronger disagreement between classifiers, Yayambo is a good fusion method for fusion of inconsistent classifiers. This indicates that it is beneficial to use Yayambo as it provides robustness to the quality (i.e. accuracy) of individual classifiers. \stodo{Repeating what was said before} %that classifiers capture 

%Also, it is observed i
Note that two individual $k$-NN classifiers in Table~\ref{tab:camaccuracy_similar} outperform all the fusion methods.
This is unsurprising for the benchmark methods.
For Yayambo, we might hope that the exchange of supports might help us avoid this, but these results illustrate that it is quite possible for classifiers with low accuracies (i.e. $h_3, h_4$ and $h_5$) to support each other's erroneous predictions, outweighing support for the correct predictions by the other classifiers (since classifiers can be wrong and in agreement): recall that the classifiers do not have a prior notion of which other classifiers perform better.
%have stronger agreement than the two classifiers with high accuracies (i.e. $h_1$ and $h_2$). So support that classifiers with low accuracies assign to each other dominate support that classifiers high accuracies assign to each other.

\subsubsection{Classifiers trained on different subsets of a data set}
The results of our proposed fusion method and the benchmark fusion methods when various classifiers are each trained on a different subset of a data set are presented in Table~\ref{tab:camaccuracy_subsets}. %Surprisingly in Table~\ref{tab:camaccuracy_subsets}, all the five classifiers give the same accuracy on the Occupancy data set. No clearer explanation to this result, it could be a coincidence.
Table~\ref{tab:different_subsets} shows the data set sizes for training of classifiers on different data sets, with data subsets randomly chosen without replacement. 

\begin{table}[h]
	% increase table row spacing, adjust to taste
	\centering
	% Some packages, such as MDW tools, offer better commands for making tables
	% than the plain LaTeX2e tabular which is used here.
	\scalebox{0.7}{
		\begin{tabular}{|l|c|c|c|c|c|c|c|}
			\hline
			Techniques&Iris&Gesture&Activity& Satellite&Digits&Occupancy&Columbia\\
			\hline
			$f_1$ &$0.6889$ &$0.3125$ &$0.6089$&$0.7065$ &$0.6420$&$0.8640$&$0.8381$ \\
			$f_2$ &$0.9111$ &$0.5268$ &$0.7728$&$0.8220$ &$0.8747$&$0.7809$&$0.5111$ \\
			$f_3$ &$0.7556$ &$0.3214$ &$0.5807$&$0.2305$ &$0.9080$&$0.7595$&$0.2032$ \\
			$f_4$ &$0.9333$ &$0.5833$ &$0.6336$&$0.7995$ &$\textbf{0.9403}$&$0.9808$&$0.5460$ \\
			$f_5$ &$0.9556$ &$0.4791$ &$0.5032$&$0.7030$ &$0.8933$&$\textbf{0.9852}$&$0.7302$ \\
			\hline
			Borda count rule &$0.9122$ &$0.6022$ &$0.7022$&$0.8001$ &$0.9014$ &$0.9234$ &$0.8015$  \\
			Majority vote rule &$0.9668$ &$0.6211$ &$0.7625$&$0.8044$ &$0.8325$&$0.9419$&$0.9143$ \\
			Product rule &$\textbf{0.9778}$ &$0.6328$ &$0.7555$&$0.8220$ &$0.9044$ &$0.9499$ &$0.8793$  \\
			Sum rule &$\textbf{0.9778}$ &$0.6429$ &$0.7644$&$0.8220$ &$0.9143$ &$0.9535$ &$0.8793$  \\
			Yayambo &$\textbf{0.9778}$ &$\textbf{0.6607}$ &$\textbf{0.7802}$&$\textbf{0.8240}$ &$0.8793$&$0.9519$&$\textbf{0.9143}$ \\
			\hline
	\end{tabular}}
	\caption[The Borda count rule, majority vote rule, product rule, sum rule and Yayambo accuracies on test data using classifiers trained on different subsets of data set.]{Accuracy values of the fusion methods on test data using classifiers trained on different subsets of data set. %Accuracy of individual classifier and fusion techniques on different subsets of data set. %Yayambo outperforms any individual classifier in this case. 
			%Bold values indicate best performance.
		 }
	\label{tab:camaccuracy_subsets}
\end{table}
%
%\begin{figure}[h]
%	\centering
%	\begin{subfigure}[h]{0.24\textwidth}
%		\includegraphics[width=\textwidth]{figs/convergence_10}
%		\caption{}
%		\label{fig:indicov31}
%	\end{subfigure}
%	\begin{subfigure}[h]{0.24\textwidth}
%		\includegraphics[width=\textwidth]{figs/convergence_25}
%		\caption{}
%		\label{fig:progresscov31}
%	\end{subfigure}
%	\begin{subfigure}[h]{0.24\textwidth}
%		\includegraphics[width=\textwidth]{figs/convergence_50}
%		\caption{}
%		\label{fig:sostraj31}
%	\end{subfigure}
%	\begin{subfigure}[h]{0.24\textwidth}
%		\includegraphics[width=\textwidth]{figs/convergence}
%		\caption{}
%		\label{fig:arstraj31}
%	\end{subfigure}
%	\caption{\small \textit{Individual coverage performance for a team of three robots.}}
%	\label{fig:experiment31}
%\end{figure}
\begin{table}[h]
	% increase table row spacing, adjust to taste
	\centering
	% Some packages, such as MDW tools, offer better commands for making tables
	% than the plain LaTeX2e tabular which is used here.
	\scalebox{0.7}{
		\begin{tabular}{|l|c|c|c|c|c|c|c|}
			\hline
			Techniques&Iris&Gesture&Activity& Satellite&Digits& Occupancy& Columbia\\
			\hline
			$f_1$ &$95$ &$544$ &$2513$&$3492$ &$9198$&$4761$&$619$ \\
			\hline
			$f_2$ &$57$ &$401$ &$3829$&$2700$ &$5723$&$7176$&$540$ \\
			\hline
			$f_3$ &$76$ &$473$ &$3563$&$3024$ &$6252$&$4271$ &$441$\\
			\hline
			$f_4$ &$95$ &$528$ &$2389$&$3898$ &$8128$&$5009$ &$620$ \\
			\hline
			$f_5$ &$56$ &$727$ &$3341$&$2875$ &$5036$&$7972$ &$517$ \\
			\hline
	\end{tabular}}
	\caption{Data set sizes for training of classifiers.}
	\label{tab:different_subsets}
\end{table}

Our fusion technique once again typically achieves accuracy better than or equal to the sum rule:
%Table~\ref{tab:camaccuracy_subsets} shows that, when classifiers are trained on different subsets of training data set, Yayambo again outperforms the sum rule on most data sets.
the approaches yield the same accuracy on one task; Yayambo outperforms the sum rule on four tasks, and the sum rule performs best in two cases.
When classifiers are trained on different data sets for the same task, we might consider them to be more likely to have differing views on the prediction of future test data points.
If this is the case, the better performance of Yayambo over the sum rule provides support to the view that Yayambo is more suitable for highly diverse classifiers, i.e. classifiers with significantly differing predictions for an observation.
%s (it depends on the extent of the differences in classifier outputs).
%On the Occupancy data set, all the classifiers and fusion methods give exactly the same accuracy.
%There may not be an explanation for the same performance of classifiers, it might be a coincidence.
%But we can comment on the same performance of our fusion methods. The classifier fusion methods achieve the same accuracy probably because they combine classifiers with exactly the same accuracy.\stodo{This is not correct.}

\begin{table}[h]
	% increase table row spacing, adjust to taste
	\centering
	% Some packages, such as MDW tools, offer better commands for making tables
	% than the plain LaTeX2e tabular which is used here.
	\scalebox{0.7}{
		\begin{tabular}{|l|c|c|c|c|c|c|c|}
			\hline
			Techniques&Iris&Gesture&Activity& Satellite&Digits&Occupancy&Columbia\\
			\hline
			$h_1$ &$\textbf{0.9778}$ &$0.6131$ &$0.7693$&$\textbf{0.824}$ &$0.8367$&$\textbf{0.9759}$&$0.1333$ \\
			$h_2$ &$0.6222$ &$0.4048$ &$0.7787$&$0.2350$ &$0.8290$&$0.9688$&$0.9428$ \\
			$h_3$ &$0.9778$ &$0.6191$ &$0.7254$&$0.2305$ &$\textbf{0.9460}$&$0.9752$&$\textbf{0.9904}$ \\
			$h_4$ &$0.2667$ &$0.4821$ &$0.8242$&$0.7810$ &$0.1000$&$0.6471$&$0.6793$ \\
			$h_5$ &$\textbf{0.9778}$ &$0.6548$ &$0.7293$&$0.8215$ &$0.1173$&$0.6010$&$0.7555$ \\
			\hline
		Borda count rule &$0.9254$ &$0.6021$ &$0.8212$&$0.8230$ &$0.9002$ &$0.9332$ &$0.9774$  \\
		Majority vote rule &$0.9262$ &$0.6033$ &$0.8112$&$0.8000$ &$0.9012$&$0.9421$&$0.9686$ \\
		Product rule &$0.9755$ &$0.6444$ &$\textbf{0.8335}$&$0.8110$ &$0.9089$ &$0.9750$ &$0.9800$  \\
			Sum rule &$\textbf{0.9778}$ &$\textbf{0.6607}$ &$\textbf{0.8335}$&$0.8230$ &$0.9177$ &$0.9752$ &$0.9841$  \\
			Yayambo &$\textbf{0.9778}$ &$0.6488$ &$0.8301$&$0.8315$ &$0.9197$&$\textbf{0.9759}$&$0.9841$ \\
			\hline
	\end{tabular}}
	\caption{Test accuracy of the benchmark fusion methods, Yayambo and individual MLP classifiers trained on different data sets. %Accuracy of individual classifier and fusion techniques on on different data sets. %Yayambo outperforms any individual classifier in this case. 
			Bold values indicate best performance. }
	\label{tab:camaccuracy_samesubsets}
\end{table}

Table~\ref{tab:camaccuracy_samesubsets} shows results of using various classifiers of the same type but with different hyperparameters (we chose MLPs in this case) when trained on different subsets of the training data sets. 
Unlike Table~\ref{tab:camaccuracy_subsets}, this table shows that when the same classifier type was trained on different subsets of a data set, Yayambo only slightly outperformed the sum rule. This might indicate that the both fusion methods are recommended when classifiers are highly similar. %This indicates that there may be situations where the sum rule could perform better than Yayambo.
%\stodo{Can we draw any conclusions from this?  Is there some kind of useful interpretation?}

\subsubsection{Classifiers trained on different subsets of features}

Table~\ref{tab:different_features} compares our proposed fusion method and the benchmark fusion methods when classifiers are trained on different features of a data set.
 The number of features considered for each data set were shown in Table~\ref{tab:data}; features were randomly selected
without replacement.

\begin{table}[h]
	% increase table row spacing, adjust to taste
	\centering
	% Some packages, such as MDW tools, offer better commands for making tables
	% than the plain LaTeX2e tabular which is used here.
	\scalebox{0.7}{
		\begin{tabular}{|l|c|c|c|c|c|c|c|}
			\hline
			Techniques&Iris&Gesture&Activity& Satellite&Digits&Occupancy&Columbia\\
			\hline
			$f_1$ &$0.2667$ &$0.3095$ &$0.2893$&$0.221$ &$0.6423$&$0.6353$&$\textbf{0.8762}$ \\
			$f_2$ &$0.6285$ &$0.6190$ &$0.7348$&$0.8535$ &$0.8846$&$\textbf{0.8671}$&$0.6095$ \\
			$f_3$ &$0.1206$ &$0.5536$ &$0.6128$&$0.8310$ &$0.8566$&$0.6352$&$0.6413$ \\
			$f_4$ &$\textbf{0.7238}$ &$\textbf{0.9048}$ &$\textbf{0.7550}$&$\textbf{0.8775}$ &$\textbf{0.9376}$&$0.5101$&$0.6413$ \\
			$f_5$ &$0.1333$ &$0.6845$ &$0.6083$&$0.8475$ &$0.893$&$0.5830$&$0.8190$ \\
			\hline
			Borda count rule &$0.1223$ &$0.4254$ &$0.2622$&$0.4880$ &$0.6910$ &$0.6155$ &$0.7587$  \\
			Majority vote rule &$0.1332$ &$0.4624$ &$0.2333$&$0.4877$ &$0.6770$ &$0.6001$ &$0.7446$  \\
			Product rule &$0.1333$ &$0.4524$ &$0.2555$&$0.4887$ &$0.6907$ &$0.6111$ &$0.7338$  \\
			Sum rule &$0.1333$ &$0.4524$ &$0.2622$&$0.4910$ &$0.6910$ &$0.6200$ &$0.7587$  \\
			Yayambo &$0.1333$ &$0.4672$ &$0.2889$&$0.5300$ &$0.6367$&$0.6000$&$0.8031$ \\
			\hline
	\end{tabular}}
	\caption{Accuracy values of the fusion methods on test data using different classifiers trained on different features of data set.  %Yayambo outperforms any individual classifier in this case. 
			%Bold values indicate best performance. 
		}
	\label{tab:different_features}
\end{table}

%\begin{table}[h]
%	% increase table row spacing, adjust to taste
%	\centering
%	% Some packages, such as MDW tools, offer better commands for making tables
%	% than the plain LaTeX2e tabular which is used here.
%	\scalebox{0.7}{
%		\begin{tabular}{|l|c|c|c|c|c|c|c|}
%			\hline
%			Data sets&Iris&Gesture&Activity& Satellite&Digits&Occupancy&Columbia\\
%			\hline
%			$n^-/n$&$2/4$ &$12/19$ &$4/6$&$25/36$ &$299/784$&$2/5$&$54/97$ \\
%			\hline
%	\end{tabular}}
%	\caption{The number of features considered to train classifiers on each data set, where $n$ indicates the total number of features from the corresponding training data set, and $n^-$ the number of features selected.}
%	\label{tab:n_features}
%\end{table}
Here  we observe that, for many of the data sets, most of the individual classifiers outperform  the fusion methods, with $f_4$ outperforming the other classifiers for most of the tasks.
We would expect some classifiers to perform poorly in some cases if the selected features do not have sufficient discriminatory information to perform high-quality classification.
However, it is unclear why $f_4$ specifically performs so well. In other words, there is no clear explanation of why fusion methods achieve poor results, a clearer explanation requires a more in-depth investigation.
%This might mean that classifiers have different behaviours for different features, and  $f_4$ works well for the selected features.
%This might indicate that when classifiers have learnt from different features, it would be desirable to randomly select and report results of a single classifier. (A classifier would need to be randomly chosen in this case because classifiers' performances on the training data sets are not available.)
The results show that our fusion technique and the sum rule method achieve roughly the same average performance (accuracy in this case) on some data sets, with no clear advantage for either approach. 

The poor performances of our fusion methods might be caused by the fact that individual classifiers were trained on different features. This indicate that our fusion methods require that individual classifiers have some regularity in the behaviours, which requires that individual classifiers be trained on the same features.
%\stodo{We really need a reasonable explanation of why the fusion results here are so poor, and mention the poor results.}

%Table~\ref{tab:different_features} shows that when classifiers are trained on different features of various data sets, Yayambo and the sum rule can perform different performance in different situations.
%For example for the Digits data set, the sum rule outperformed Yayambo. On the Columbia data set, results indicate that Yayambo achieved better performance than the sum rule.
%This might indicate that, when classifiers have been trained on different features of the same observations, there is no clear preference for Yayambo over the sum rule. 

\subsubsection{Increasing the number of classifiers}
Table~\ref{tab:progressive_fusion} shows the accuracies of the fusion methods when increasing the number $m$ of classifiers.
For the selection of classifiers, we sample the number of classifiers to be considered in the range $2\leq m\leq 5$: we sample $3$ classifiers $3$ times and average their performances. %st $m$ classifiers, which means that if $m=3$, we will consider $f_1, f_2$ and $f_3$. But one can apply another way of selecting classifiers. 
For two classifiers, the sum rule generally outperformed Yayambo.
However, for $m\geq 3$, Yayambo largely outperformed the sum rule.
(On the Iris and Occupancy data sets, both techniques yield almost identical results.
This is probably because the classifiers trained on these two data sets have fairly similar behaviour, as evidenced by their similar accuracies.)
%Table \ref{tab:progressive_fusion} shows results on the fusion methods when we consider an increasing number of classifiers. When we consider two classifiers, both fusion methods yield almost the same accuracy (the sum rule does better in some cases). But as the number of classifiers considered increases, Yayambo seems to achieve better than the sum rule in general, particularly for Gesture, Activity, Satellite, Digits and Columbia data set. On the Iris and Occupancy data sets, the both techniques yield similar results. This is probably because classifiers trained on these two data sets are fairly consistent, i.e. they give almost the same accuracy.
%Obviously, the number of classifiers can influence the quality of the fused output. The quality of the fusion output in this case will also depend on the quality of each classifier. 
%To conclude this section,
%Besided finding Yayambo to be more beneficial with more classifiers, we also observed that it was more robust to one poor classifier.
%\stodo{No results reported here support this further observation - expland?}

\begin{table}[h]
	% increase table row spacing, adjust to taste
	\centering
	% Some packages, such as MDW tools, offer better commands for making tables
	% than the plain LaTeX2e tabular which is used here.
	\scalebox{0.6}{
		\begin{tabular}{ |c|l|c|c|c|c|c|c|c| }
			\hline
			\#Classifiers&Methods&Iris&Gesture& Activity&Satellite&Digits&Occupancy&Columbia\\
			\hline
			\multirow{2}{*}{$2$} & Borda count rule & $\textbf{0.9111}$ & $\textbf{0.5514}$ &$0.7444$&$0.8005$&$0.8651$&$\textbf{0.6352}$&$0.5212$\\
			& Majority vote rule & $\textbf{0.9111}$ & $\textbf{0.5514}$ &$0.7364$&$0.76604$&$0.8575$&$0.6222$&$\textbf{0.5333}$ \\
			& Product rule & $\textbf{0.9111}$ & $\textbf{0.5514}$ &$0.7444$&$\textbf{0.8225}$&$0.8762$&$\textbf{0.6352}$&$\textbf{0.5333}$\\
			& Sum rule & $\textbf{0.9111}$ & $\textbf{0.5514}$ &$\textbf{0.7501}$&$\textbf{0.8225}$&$\textbf{0.8771}$&$\textbf{0.6352}$&$\textbf{0.5333}$\\
			& Yayambo & $\textbf{0.9111}$ & $\textbf{0.5514}$ &$0.7466$&$0.7801$&$\textbf{0.8771}$&$\textbf{0.6352}$&$\textbf{0.5333}$ \\
			\hline
			\multirow{2}{*}{$3$} & Borda count rule & $0.9777$ & $0.5465$ &$0.7446$&$0.8032$&$0.9230$&$0.6310$&$ 0.7596$\\
			& Majority vote rule & $\textbf{0.9778}$ & $0.5449$ &$0.7822$&$0.8020$&$0.9111$&$0.6223$&$\textbf{0.8889}$ \\
		& Product rule & $\textbf{0.9778}$ & $0.5444$ &$0.7348$&$0.8060$&$0.9207$&$0.6298$&$ 0.7685$\\
		& Sum rule & $\textbf{0.9778}$ & $0.5476$ &$0.7536$&$0.8130$&$0.9207$&$\textbf{0.6352}$&$ 0.7685$\\
		& Yayambo & $\textbf{0.9778}$ & $\textbf{0.5595}$ &$\textbf{0.7837}$&$\textbf{0.8150}$&$\textbf{0.9257}$&$\textbf{0.6352}$&$\textbf{0.8889}$ \\
			\hline
			\multirow{2}{*}{$4$} & Borda count rule & $0.9556$ & $0.6339$ &$0.7544$&$0.8115$&$0.9229$&$0.6345$&$0.7443$\\
			& Majority vote rule & $0.9556$ & $0.6499$ &$0.7900$&$0.8200$&$0.9334$&$\textbf{0.6407}$&$0.8777$ \\
			& Product rule & $0.9556$ & $0.6429$ &$0.7646$&$\textbf{0.8265}$&$0.9322$&$0.625$&$0.7522$\\
			& Sum rule & $0.9556$ & $0.6429$ &$0.7644$&$\textbf{0.8265}$&$0.9393$&$0.6353$&$0.7524$\\
			& Yayambo & $\textbf{0.9778}$ & $\textbf{0.6458}$ &$\textbf{0.7906}$&$0.8230$&$\textbf{0.9433}$&$\textbf{0.6407}$&$\textbf{0.8857}$ \\
			\hline
			\multirow{2}{*}{$5$} & Borda count rule&$0.9777$ &$0.6455$ &$0.7356$&$0.8090$ &$0.8905$ &$0.9033$ &$0.8878$  \\
			&Majority vote rule&$\textbf{1.0000}$ &$0.6728$ &$0.7446$&$0.8122$ &$0.8886$ &$0.9133$ &$0.8996$  \\
			&Product rule &$\textbf{1.0000}$ &$0.6665$ &$0.7455$&$0.8110$ &$0.8905$ &$0.9042$ &$0.8852$  \\
			&Sum rule &$\textbf{1.0000}$ &$0.6875$ &$0.7644$&$0.8010$ &$0.9153$ &$0.9163$ &$0.9048$  \\
			&Yayambo &$\textbf{1.0000}$ &$\textbf{0.7054}$ &$\textbf{0.7788}$&$\textbf{0.8120}$ &$\textbf{0.9167}$&$\textbf{0.9365}$&$\textbf{0.9535}$ \\
			\hline
	\end{tabular}}
	\caption[The Borda count rule, majority vote rule, product rule, sum rule and Yayambo accuracies on test data.]{Accuracy values of fusion methods on benchmark data sets. %The value $m$ denotes the number of classifiers considered for each case. %Sum rule outperforms Yayambo. Individual classifiers have low accuracies in four datasets.
			%Bold values indicate best performance.
		}
	\label{tab:progressive_fusion}
\end{table}

\subsubsection{Artificial classifiers}

Our empirical results so far provide evidence that the Yayambo fusion approach often outperforms the Borda count rule, the majority vote rule, the product rule and the sum rule, but we also see cases where this does not hold.
Here we consider fusing outputs of some hypothetical classifiers which we define by specifying their behaviour, rather than training them. 
The hope is that the extreme setting we describe here sheds further light on the behaviour of our proposed fusion method, for which results are presented in Tables \ref{tab:disagreement0} (where we consider an increasing number of classifiers) and \ref{tab:disagreement1} (where we consider all classifiers).

For the artificial classifier outputs, we have the following situation.
We consider fusing five classifiers $f_1$ to $f_5$ for a binary classification task, where:
\begin{itemize}
	\item $f_1$ always classifies correctly, with probability predictions $0.51$ for the correct class and $0.49$ for the incorrect class;
	\item $f_2$ also classifies perfectly, but it assigns probabilities $0.9$ and $0.1$ for the correct and incorrect class respectively;
	\item $f_3$ is always wrong: it predicts $0.49$ for the correct class, and $0.51$ for the incorrect class;
	\item $f_4$ correctly classifies a test point with an assigned probability of $0.7$ in $65\%$ of cases, and in the other 35\% outputs a distribution with a first class probability sampled uniformly from $(0,1)$; and
	%	%over the two classes sampled from the Dirichlet distribution
		%with concentration parameter $\beta=1$; and
		%\stodo{Beta distribution?  Actually a Bernoulli distribution with parameter from the Beta distribution.}
		%($\beta$ can be interpreted as a pseudocount of successes and failures in Dirichlet distribution);
	\item $f_5$ always outputs a distribution with a first class probability sampled uniformly from $(0,1)$.
	%	over the two classes sampled from the Dirichlet distribution with $\beta = 1$.
\end{itemize}
To be able to evaluate accuracies of our artificial classifiers, we generate $5000$ expected predictions for each class. Expected performance for each of $5$ individual artificial classifiers is shown on the second column of Table \ref{tab:disagreement0}.
%
%

%(Tables \ref{tab:disagreement0} and \ref{tab:disagreement1}) 
%\stodo{Add text to refer to tables...}

Note that viewed at the output level, the first and the third classifiers are in close agreement---see the small value at the intersection of $f_1$ and $f_3$ in Table \ref{tab:disagreement_pred}---while they differ substantially from the other three classifiers. 
On the other hand, from a decision perspective, the first and the second classifiers agree exactly---see the high value at the intersection of $f_1$ and $f_2$ in Table \ref{tab:disagreement_acc}; we used accuracy to evaluate classification disagreement between two classifiers, while the third classifier disagrees totally with the first two classifiers. The fourth and fifth classifiers can agree or disagree with others depending on the random outputs.

%\begin{table}[h]
%	% increase table row spacing, adjust to taste
%	\centering
%	% Some packages, such as MDW tools, offer better commands for making tables
%	% than the plain LaTeX2e tabular which is used here.
%	\scalebox{0.7}{
%		\begin{tabular}{|l|c|}
%			\hline
%			Techniques&Expected accuracy\\
%			\hline
%			$f_1$ & $1.0000$  \\
%			$f_2$ & $1.0000$  \\
%			$f_3$ & $0.0000$ \\
%			$f_4$ & $0.8250$  \\
%			$f_5$ & $0.5000$ \\
%			\hline
%	\end{tabular}}
%	\caption[\textit{Expected performances of individual artificial classifiers.}]{Expected performances of individual artificial classifiers.}
%	\label{tab:expected_performance}
%\end{table}

\begin{table}[h]
	% increase table row spacing, adjust to taste
	\centering
	% Some packages, such as MDW tools, offer better commands for making tables
	% than the plain LaTeX2e tabular which is used here.
	\scalebox{0.7}{
		\begin{tabular}{|l|c|c|c|c|c|}
	\hline
	Techniques& Expected accuracy & Case 1&Case 2&Case 3& Case 4\\
	\hline
	$f_1$ & $1.0000$  &$\textbf{1.0000}$ &$\textbf{1.0000}$ &$\textbf{1.0000}$&$\textbf{1.0000}$ \\
	$f_2$ & $1.0000$ &$\textbf{1.0000}$ &$\textbf{1.0000}$ &$\textbf{1.0000}$&$\textbf{1.0000}$ \\
	$f_3$ &$0.0000$ &$-$ &$0.0000$ &$0.0000$&$0.0000$ \\
	$f_4$ & $0.8250$  &$-$ &$-$ &$0.8242$&$0.8242$ \\
	$f_5$ & $0.5000$  &$-$ &$-$ &$-$&$0.5017$ \\
	\hline
	Borda count rule & $-$ &  $\textbf{1.0000}$ &$\textbf{1.0000}$ &$0.9123$&$0.9123$ \\
	Majority vote rule& $-$ &$\textbf{1.0000}$ &$\textbf{1.0000}$ &$0.9123$&$0.9123$ \\
	Product rule& $-$  &$\textbf{1.0000}$ &$\textbf{1.0000}$ &$0.9617$&$0.9011$ \\
	Sum rule& $-$ &$\textbf{1.0000}$ &$\textbf{1.0000}$ &$0.9617$&$0.9365$ \\
	Yayambo & $-$ &$\textbf{1.0000}$ &$\textbf{1.0000}$ &$0.9617$&$0.9333$ \\
	\hline
\end{tabular}}
	\caption[The Borda count rule, majority vote rule, product rule, sum rule and Yayambo accuracies using artificial classifiers.]{Accuracy values of the fusion methods using artificial classifiers. Here, cases refer to numbers of classifiers considered. %Yayambo outperforms any individual classifier in this case. 
		%	Bold values indicate best performance.
		 }
	\label{tab:disagreement0}
\end{table}

\begin{table}[h]
	% increase table row spacing, adjust to taste
	\centering
	% Some packages, such as MDW tools, offer better commands for making tables
	% than the plain LaTeX2e tabular which is used here.
	\scalebox{0.7}{
		\begin{tabular}{|l|c|c|c|c|}
	\hline
	Techniques&Case 1&Case 2&Case 3\\
	\hline
	$f_1$ &$\textbf{1.0000}$ &$\textbf{1.0000}$ &$\textbf{1.0000}$ \\
	$f_2$ &$0.0000$ &$\textbf{1.0000}$ &$\textbf{1.0000}$ \\
	$f_3$ &$0.0000$ &$0.0000$ &$0.0000$ \\
	$f_4$ &$0.8221$ &$0.8280$ &$0.8164$ \\
	$f_5$ &$0.5013$ &$0.5088$ &$0.5050$\\
	\hline
	Borda count rule &$0.9099$ &$0.9137$ &$0.9074$\\
	Majority vote rule &$0.9087$ &$0.9097$ &$0.9017$ \\
	Product rule &$0.9087$ &$0.9097$ &$0.9017$\\
	Sum rule &$0.9376$ &$0.9387$ &$0.9283$\\
	Yayambo &$0.9350$ &$0.9344$ &$0.9253$ \\
	\hline
\end{tabular}}
	\caption[\textit{The Borda count rule, majority vote rule, product rule, sum rule and Yayambo accuracies using artificial classifiers for different runs of the fourth and fifth classifiers as they are purely random.}]{Accuracy values of the fusion methods using artificial classifiers. %Yayambo outperforms any individual classifier in this case. 
			Here, cases refer to different runs of the fourth and fifth classifiers as they are purely random.
			%Bold values indicate best performance.
		 }
	\label{tab:disagreement1}
\end{table}

%For the second situation (Figure \ref{fig:disagreement}), we consider $5$ random classifiers and 30 simulated runs. We have $30$ simulated runs where in each $5$ random classifiers generate their probability distribution over classes (we took $5$ classes) following the Dirichlet distribution with $\beta = 1$. In this setting, classifiers disagree substantially from one another.
%
%\begin{figure}[h]
%	\centering
%	\begin{subfigure}[h*]{0.48\textwidth}
%		\includegraphics[width=\textwidth]{figs/classifier/disagreement_0.png}
%		\caption{}
%		\label{fig:disagreement0}
%	\end{subfigure}
%	\begin{subfigure}[h*]{0.48\textwidth}
%		\includegraphics[width=\textwidth]{figs/classifier/disagreement_1.png}
%		\caption{}
%		\label{fig:disagreement1}
%	\end{subfigure}
%	\caption[\textit{Plots of $30$ simulated runs using the sum rule method and Yayambo on the first artificial data set.}]{ \textit{Plots of $30$ simulated runs using the sum rule method and Yayambo on the first artificial data set. \textbf{(\ref{fig:disagreement0})}: Accuracies of sum rule, Yayambo and the best classifier. \textbf{(\ref{fig:disagreement1})}: Accuracies of sum rule and Yayambo. %The normalised coverage (in $\%$) intervals obtained with SOS, ARS and PRS are $83.140\pm 6.913$, $68.108\pm 13.189$ and $37.016\pm 4.269$ for the eight robots respectively; and $80.239\pm 5.909$, $63.053\pm 13.046$ and $31.090\pm 2.574$ for the ten robots respectively.
%	}}
%	\label{fig:disagreement}
%\end{figure}

\begin{table}[h]
	% increase table row spacing, adjust to taste
	\centering
	% Some packages, such as MDW tools, offer better commands for making tables
	% than the plain LaTeX2e tabular which is used here.
	\scalebox{0.7}{
		\begin{tabular}{|l|c|c|c|c|}
	\hline
	&$f_2$&$f_3$&$f_4$&$f_5$\\
	\hline
	$f_1$ &$55.1443$ &$2.8284$&$32.66384$ &$41.0806$\\
	$f_2$ &$-$ &$57.9228$&$48.1349$&$69.7933$ \\
	$f_3$ &$-$ &$-$&$34.1587$&$41.0897$ \\
	$f_4$ &$-$ &$-$&$-$&$51.7427$ \\
	\hline
\end{tabular}}
	\caption[Measurement of disagreement between classifiers.]{Measurement of disagreement between classifiers from a prediction point of view. %Yayambo outperforms any individual classifier in this case. 
			We used the Euclidean distance between classifiers' outputs to evaluate prediction disagreement between classifiers.
			Values range in $[0, \infty)$ and high values indicate low agreement between classifiers based on classifiers' predictions. }
	\label{tab:disagreement_pred}
\end{table}
\balance
\begin{table}[h]
	% increase table row spacing, adjust to taste
	\centering
	% Some packages, such as MDW tools, offer better commands for making tables
	% than the plain LaTeX2e tabular which is used here.
	\scalebox{0.7}{
		\begin{tabular}{|l|c|c|c|c|}
	\hline
	&$f_2$&$f_3$&$f_4$&$f_5$\\
	\hline
	$f_1$ &$1.0000$ &$0.0000$&$0.8164$ &$0.5050$\\
	$f_2$ &$-$ &$0.0000$&$0.8164$&$0.5050$ \\
	$f_3$ &$-$ &$-$&$0.1836$&$0.4950$ \\
	$f_4$ &$-$ &$-$&$-$&$0.5066$ \\
	\hline
\end{tabular}}
	\caption[Measurement of agreement between classifiers]{Measurement of agreement between two classifiers from classification point of view using accuracy where the first classifier's outputs are considered as ground truth and the second classifier's outputs as predicted outputs. %Yayambo outperforms any individual classifier in this case. 
			Values range in $[0, 1]$ and high values indicate high agreement between classifiers based on their accuracy. }
	\label{tab:disagreement_acc}
\end{table}
%\section{Discussion}
%\label{sec:fusion_discussion}
%\subsection{Classifiers trained on the same data set}

%\subsection{Similar classifiers trained with different parameters on the same data set}

%\subsection{Classifiers trained on different subsets of a data set}

%\subsection{Classifiers trained on different subsets of features}

%\subsection{On increasing the number of classifiers}

%\subsection{A case study for active vision}

%\subsection{Classifiers trained on artificial data set}
%
%This claim could also be applied to the sum rule method.

%It is observed that the best classifier achieves higher accuracy than both fusion methods.
%In particular, 
Table~\ref{tab:disagreement0} confirms that the accuracy of fused decisions may degrade as we add weaker (or erratic) classifiers.
These results support our earlier argument that Yayambo relies on regularity in the behaviour of classifiers resulting from their being trained on the same task, resulting in similar behaviour on future observations.
%for it to outperform individual classifiers. %But these artificial classifiers do not have such regularity, because they were not trained.
%\stodo{Can we say anything further that is useful here?}

\section{Conclusions and future work}
\label{chp:conclusion}
%}

We proposed a classifier fusion method which combines classifiers' outputs, even though knowledge about their functioning or prior performance is not available.
The method attempts to take advantage of the expectation of similar behaviour of classifiers trained for the same task.
The output of the method is focused on consensus-driven decision-making, where the loss function is symmetric.
Since the method does not aim to output calibrated probabilities for the various classes, it is not recommended for use in downstream tasks where such probabilities are desirable.

%Our proposed method is centralised,  which means that  there exists a combiner which collects initial distributions (i.e. classifiers' outputs) and combines them.   %Details of our contributions and novelties for classifier fusion can be found in Section \ref{sec:network_contribution}.

Our experiments compare our approach to four established black-box classifier fusion approaches, the Borda count rule, the majority vote rule, the product rule and the sum rule.
%Furthermore, 
We found that our proposed method generally outperformed these fusion approaches, yielding the best accuracy in many of the cases we considered.
This observation held for both highly similar and highly diverse classifiers, indicating that our proposed method is more robust to disparities in the quality of individual classifiers than the sum rule, the product rule, the majority vote rule or the Borda count rule.
%Since our proposed method is not a probabilistic method, which results a high prediction loss. So Yayambo is not suitable in situations where the misclassification loss function is  appropriate. %Our classifier fusion method can give highly uncertain predictions, but which are highly accurate. 
%We believe that this conflict between loss and accuracy is because our proposed method assigns high probability (of almost one) to the consensus class. So when there is misclassification, the loss is high. 
%We analyse the conflict between loss and accuracy \cite{conflict2018} observed in Yayambo next.

It should be noted that the sum rule and product rule are recommended for generating calibrated fused outputs; it is possible that still other fusion approaches might be recommended for consensus-based decision-making.
In our experiments we observed empirically that even though supports assigned to predictions are asymmetric, the updated probability vectors converge. 
%This is interesting as from all the probability vectors we can output the same class of an object of interest for which classifier outputs confer differing views before.
While a consensus decision is not necessarily correct, we contend that achieving consensus from different initial distributions using such an asymmetric notion of support confers some credence on the final decision.

There are a number of avenues of interesting theoretical work to further develop our understanding of the Yayambo fusion method.
First, it would be good to establish convergence theoretically.
Further, it may be possible to obtain a closed form expression for the consensus class from the initial probabilities without performing the iteration explicitly, possibly with some modifications to the forms of the equations used for the supports.
Finally, it would be interesting to consider how to formalize a probabilistic prior interpretation of the regularity assumption we are making, and what posterior it leads to in the Bayesian probability setting.
The connections this could lead to may lead to support formulae with better theoretical motivations and further improved performance.
It would also lay a solid foundation for fusion of multiple sequential observations, rather than the once-off case considered by our algorithm.

\vskip 0.2in
\bibliographystyle{unsrt}
% argument is your BibTeX string definitions and bibliography database(s)
\bibliography{references}

\end{document}